\documentclass[10pt,twocolumn,letterpaper]{article}

\usepackage[pagenumbers]{cvpr} %

\usepackage{graphicx}
\usepackage{amsmath}
\usepackage{amssymb}
\usepackage{booktabs}
\usepackage{algorithm,algorithmicx,algpseudocode}
\usepackage{mathtools}
\usepackage{amssymb}
\usepackage{xcolor}

\usepackage[accsupp]{axessibility}

\usepackage{pythonhighlight}

\usepackage[pagebackref,breaklinks,colorlinks]{hyperref}

\usepackage{array}
\newcolumntype{C}[1]{>{\centering\let\newline\\\arraybackslash\hspace{0pt}}m{#1}}
\newcommand{\parsection}[1]{\vspace{0.5mm}\noindent\textbf{#1:}~}

\newcommand{\captionRow}[2]{
    \resizebox{\linewidth}{!}{
        \begin{tabular}{ C{#1cm} C{#1cm} C{#1cm} C{#1cm} C{#1cm} C{#1cm} C{#1cm} C{#1cm} C{#1cm} C{#1cm} C{#1cm} C{#1cm} C{#1cm} C{#1cm} C{#1cm} C{#1cm} }
        #2
    \end{tabular}
    }}

\newcommand{\captionRowEight}[2]{
    \resizebox{\linewidth}{!}{
        \begin{tabular}{ C{#1cm} C{#1cm} C{#1cm} C{#1cm} C{#1cm} C{#1cm} C{#1cm} C{#1cm} }
        #2
    \end{tabular}
    }}
    
\newcommand{\captionRowSeven}[2]{
    \resizebox{\linewidth}{!}{
        \begin{tabular}{ C{#1cm} C{#1cm} C{#1cm} C{#1cm} C{#1cm} C{#1cm} C{#1cm} }
        #2
    \end{tabular}
    }}
    
\newcommand{\captionRowSix}[2]{
    \resizebox{\linewidth}{!}{
        \begin{tabular}{ C{#1cm} C{#1cm} C{#1cm} C{#1cm} C{#1cm} C{#1cm} }
        #2
    \end{tabular}
    }}

\usepackage{calc}

\usepackage[capitalize]{cleveref}
\crefname{section}{Sec.}{Secs.}
\Crefname{section}{Section}{Sections}
\Crefname{table}{Table}{Tables}
\crefname{table}{Tab.}{Tabs.}

\def\cvprPaperID{5974} %
\def\confName{CVPR}
\def\confYear{2022}

\usepackage{capt-of,etoolbox}
\makeatletter
\apptocmd\@maketitle{{\introfig{}\par}}{}{}
\makeatother

\newcommand{\E}{\mathbb{E}}

\newcommand{\Ea}[1]{\E\left[#1\right]}

\newcommand{\kl}[2]{D_{\mathrm{KL}}\!\left(#1 ~ \| ~ #2\right)}

\newcommand{\bI}{\mathbf{I}}

\newcommand{\bzero}{\mathbf{0}}

\newcommand{\bx}{\mathbf{x}}

\newcommand{\bz}{\mathbf{z}}

\newcommand{\bepsilon}{{\boldsymbol{\epsilon}}}

\usepackage{fontawesome}

\usepackage{fancyhdr}
\fancypagestyle{plain}{%
  
  \lhead{\small{{\color{gray}Extended Version of accepted CVPR 2022 Paper}}}
  \rhead{}
}
\begin{document}

\newcommand{\introfig}{\protect\centering\vspace{-5mm}
\includegraphics[width=\linewidth]{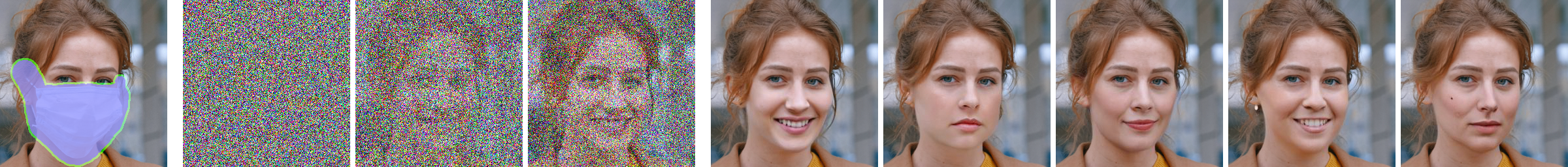} 
\includegraphics[width=\linewidth]{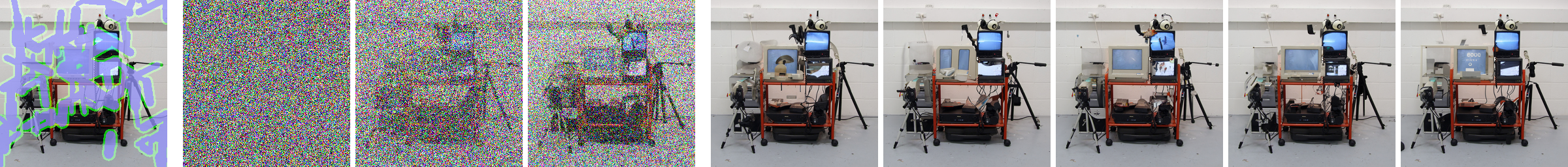}
\includegraphics[width=\linewidth]{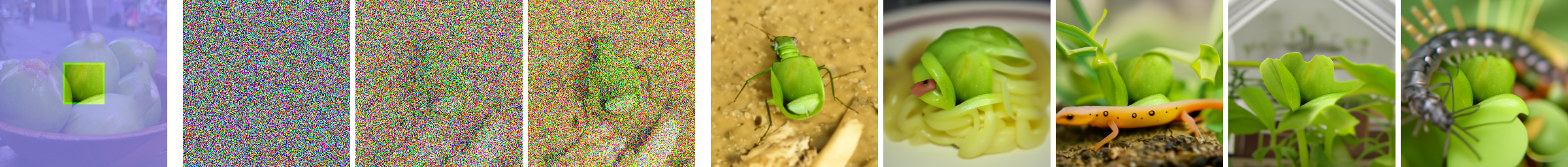}
\includegraphics[width=\linewidth]{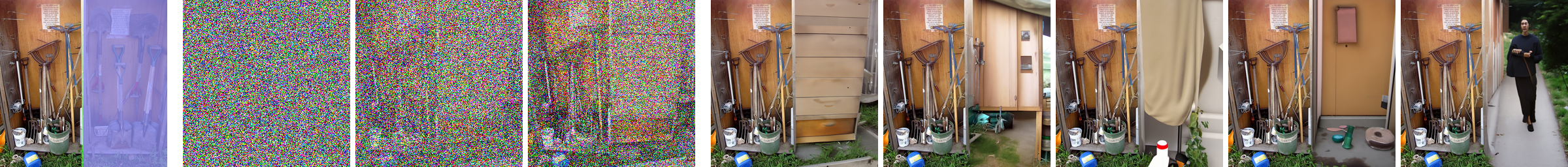}
\resizebox{\linewidth}{!}{
    \begin{tabular}{ C{2.5cm} C{2.5cm} C{2.5cm} C{2.5cm} C{0cm} C{2.5cm} C{2.5cm} C{2.5cm} C{2.5cm} C{2.5cm} }
    Input &
    ~~~Denoising 0\%  &
    ~~~Denoising 60\% &
    ~~~Denoising 75\% &
    \hfill &
    Sample 1 &
    Sample 2 &
    Sample 3 &
    Sample 4 &
    Sample 5
\end{tabular}}
\captionof{figure}{\textbf{We use Denoising Diffusion Probabilistic Models (DDPM) for inpainting.}
The process is conditioned on the masked input (\textit{left}). It starts from a random Gaussian noise sample that is iteratively denoised until it produces a high-quality output.
Since this process is stochastic, we can sample multiple diverse outputs.
The DDPM prior forces a harmonized image, is able to reproduce texture from 
other regions, and inpaint semantically meaningful content.}%
\label{fig:intro}
\vspace{9mm}
}

\title{RePaint: Inpainting using Denoising Diffusion Probabilistic Models}

\newcommand{\aand}{\hspace{3mm}}
\author{Andreas Lugmayr \aand Martin Danelljan \aand Andres Romero \aand Fisher Yu \aand Radu Timofte \aand Luc Van Gool \vspace{3mm}\\
Computer Vision Lab\vspace{3mm} \\
ETH Z\"urich, Switzerland
}

\maketitle

\begin{abstract}
Free-form inpainting is the task of adding new content to an image in the regions specified by an arbitrary binary mask.
Most existing approaches train for a certain distribution of masks, which limits their generalization capabilities to unseen mask types. 
Furthermore, training with pixel-wise and perceptual losses often leads to simple textural extensions towards the missing areas instead of semantically meaningful generation.
In this work, we propose RePaint: A Denoising Diffusion Probabilistic Model (DDPM) based inpainting approach that is applicable to even extreme masks.
We employ a pretrained \emph{unconditional} DDPM as the generative prior. To condition the generation process, we only alter the reverse diffusion iterations by sampling the unmasked regions using the given image information.
Since this technique does not modify or condition the original DDPM network itself, the model produces high-quality and diverse output images for any inpainting form.
We validate our method for both faces and general-purpose image inpainting using standard and extreme masks.
RePaint outperforms state-of-the-art Autoregressive, and GAN approaches for at least five out of six mask distributions.
Github Repository: \textcolor{blue}{\href{https://www.git.io/RePaint}{git.io/RePaint}}
\vspace{-0mm}

\end{abstract}

\section{Introduction}
\label{sec:intro}

Image Inpainting, also known as Image Completion, aims at filling missing regions within an image.
Such inpainted regions need to harmonize with the rest of the image and be semantically reasonable. Inpainting approaches thus require strong generative capabilities.
To this end, current State-of-the-Art approaches~\cite{yu2019free,liu2020rethinking,aot,lama} rely on GANs~\cite{goodfellow2014generative} or Autoregressive Modeling~\cite{dsi,ict,yu2021diverse}.
Moreover, inpainting methods need to handle various forms of masks such as thin or thick brushes, squares, or even extreme masks where the vast majority of the image is missing. This is highly challenging since existing approaches train with a certain mask distribution, which can lead to poor generalization to novel mask types. In this work, we investigate an alternative generative approach for inpainting, aiming to design an approach that requires no mask-specific training.

Denoising Diffusion Probabilistic Models (DDPM) is an emerging alternative paradigm for generative modelling~\cite{ddpm,diffusionThermo}. Recently, Dhariwal and Nichol~\cite{beatGan} demonstrated that DDPM can even outperform the state-of-the-art GAN-based method~\cite{brock2018large} for image synthesis. 
In essence, the DDPM is trained to iteratively denoise the image by reversing a diffusion process.
Starting from randomly sampled noise, the DDPM is then iteratively applied for a certain number of steps, which yields the final image sample.
While founded in principled probabilistic modeling, DDPMs have been shown to generate diverse and high-quality images~\cite{ddpm,improvedddpm,beatGan}.

We propose RePaint: an inpainting method that solely leverages an off-the-shelf unconditionally trained DDPM. Specifically, instead of learning a mask-conditional generative model, we condition the generation process by sampling from the given pixels during the reverse diffusion iterations. Remarkably, our model is therefore not trained for the inpainting task itself. This has two important advantages.
First, it allows our network to generalize to any mask during inference.
Second, it enables our network to learn more semantic generation capabilities since it has a powerful DDPM image synthesis prior (Figure~\ref{fig:intro}).

Although the standard DDPM sampling strategy produces matching textures, the inpainting is often semantically incorrect.
Therefore, we introduce an improved denoising strategy that \textit{resamples} (RePaint) iterations to better condition the image.
Notably, instead of slowing down the diffusion process~\cite{beatGan}, our approach goes forward and backward in diffusion time, producing remarkable semantically meaningful images.
Our approach allows the network to effectively harmonize the generated image information during the entire inference process, leading to a more effective conditioning on the given image information.

We perform experiments on CelebA-HQ~\cite{celeba} and ImageNet~\cite{imagenet}, and compare with other State-of-the-Art inpainting approaches.
Our approach generalizes better and has overall more semantically meaningful inpainted regions.

\section{Related Work}
Early attempts on Image Inpainting or Image Completion exploited low-level cues within the input image~\cite{bertalmio2000image,ballester2001filling,bertalmio2003simultaneous}, or within the neighbor of a large image dataset~\cite{hays2007scene} to fill the missing region. 

\parsection{Deterministic Image Inpainting}
Since the introduction of GANs~\cite{goodfellow2014generative}, most of the existing methods follow a standard configuration, first proposed by Pathak~\etal~\cite{pathak2016context}, that is, using an encoder-decoder architecture as the main inpainting generator, adversarial training, and tailored losses that aim at photo-realism. Follow-up works have produced impressive results in recent years~\cite{ren2019structureflow,zeng2019learning,liu2020rethinking,ntavelis2020aim,hui2020image}.

As image inpainting requires a high-level semantic context, and to explicitly include it in the generation pipeline, there exist hand-crafted architectural designs such as Dilated Convolutions~\cite{iizuka2017globally,yu2015multi} to increase the receptive field, Partial Convolutions~\cite{liu2018image} and Gated Convolutions~\cite{yu2019free} to guide the convolution kernel according to the inpainted mask, Contextual Attention~\cite{yu2018generative} to leverage on global information, Edges maps~\cite{nazeri2019edgeconnect,xiong2019foreground,xu2020e2i,guo2021image} or Semantic Segmentation maps~\cite{hong2018learning,ntavelis2020sesame} to further guide the generation, and Fourier Convolutions~\cite{lama} to include both global and local information efficiently. Although recent works produce photo-realistic results, GANs are well known for textural synthesis, so these methods shine on background completion or removing objects, which require repetitive structural synthesis, and struggle with semantic synthesis (Figure~\ref{fig:sota_inet_all}).

\parsection{Diverse Image Inpainting}
Most GAN-based Image Inpainting methods are prone to deterministic transformations due to the lack of control during the image synthesis. To overcome this issue, Zheng~\etal~\cite{zheng2019pluralistic} and Zhao~\etal~\cite{zhao2020uctgan} propose a VAE-based network that trade-offs between diversity and reconstruction. Zhao~\etal~\cite{zhao2021large}, inspired by the StyleGAN2~\cite{karras2020analyzing} modulated convolutions, introduces a co-modulation layer for the inpainting task in order to improve both diversity and reconstruction.
A new family of auto-regressive methods~\cite{yu2021diverse,dsi,ict}, which can handle irregular masks, has recently emerged as a powerful alternative for free-form image inpainting.

\parsection{Usage of Image Prior}
In a different direction closer to ours Richardson~\etal~\cite{richardson2021encoding} exploits the StyleGAN~\cite{karras2019style} prior to successfully inpaint missing regions.
However, similar to super-resolution methods~\cite{menon2020pulse, chan2021glean} that leverage the StyleGAN latent space, it is to limited specific scenarios like faces.
Noteworthy, a Ulyanov~\etal~\cite{deepImagePrior} showed that the structure of a non-trained generator network contains an inherent prior that can be used for inpaining and other applications.
In contrast to these methods, we are leveraging on the high expressiveness of a pretrained Denoising Diffusion Probabilistic Model~\cite{ddpm} (DDPM) and therefore use it as a prior for generic image inpainting. Our method generates very detailed, high-quality images for both semantically meaningful generation and texture synthesis. Moreover, our method is not trained for the image inpainting task, and instead, we take full advantage of the prior DDPM, so each image is optimized independently.

\parsection{Image Conditional Diffusion Models}
The work by Sohl-Dickstein~\etal~\cite{diffusionThermo} applied early diffusion models to inpainting. More recently, Song~\etal~\cite{sde} develop a score-based formulation using stochastic differential equations for unconditional image generation, with an additional application to inpainting. However, both these works only show qualitative results, and do not compare with other inpainting approaches. In contrast, we aim to advance the state-of-the-art in image inpainting, and provide comprehensive comparisons with the top competing methods in literature.

A different line of research is guided image synthesis with DDPM-based approaches~\cite{ilvr,sdedit}.
In the case of ILVR~\cite{ilvr}, a trained diffusion model is guided using the low-frequency information from a  conditional image.
However, this conditioning strategy cannot be adopted for inpainting, since both high and low-frequency information is absent in the masked-out regions.
Another approach for image-conditional synthesis is developed by~\cite{sdedit}.
Guided generation is performed by initializing the reverse diffusion process from the guiding image at some intermediate diffusion time.
An iterative strategy, repeating the reverse process several times, is further adopted to improve harmonization.
Since a guiding image is required to start the reverse process at an intermediate time step, this approach is not applicable to inpainting, where new image content needs to be generated solely conditioned on the non-masked pixels.
Furthermore, the resampling strategy proposed in this work differs from the concurrent~\cite{sdedit}. 
We proceed through the full reverse diffusion process, starting at the end time, at each step jumping back and forth a fixed number of time steps to progressively improve generation quality.

While we propose a method that conditions an unconditionally trained model, the concurrent work~\cite{glide} is based on classifier-free guidance~\cite{ho2021classifier} for training an image-conditional diffusion model.
Another direction for image manipulation is image-to-image translation using diffusion models as explored in the concurrent work~\cite{palette}.
It trains an image-conditional DDPM, and shows an application to inpainting.
Unlike both these concurrent works, we leverage an unconditional DDPM and only condition through the reverse diffusion process itself. It allows our approach to effortlessly generalize to any mask shape for free-form inpainting. Moreover, we propose a sampling schedule for the reverse process, which greatly improves image quality.

\section{Preliminaries: Denoising Diffusion Probabilistic Models}
\label{sec:background}

In this paper, we use diffusion models~\cite{diffusionThermo} as a generative method.
As other generative models, the DDPM learns a distribution of images given a training set.
The inference process works by sampling a random noise vector ${x_T}$ and gradually denoising it until it reaches a high-quality output image ${x_0}$.
During training, DDPM methods define a diffusion process that transforms an image $x_0$ to white Gaussian noise $x_T \sim \mathcal{N}(0,1)$ in $T$ time steps. Each step in the forward direction is given by,
\begin{alignat}{2}
    q(x_t | x_{t-1}) & = \mathcal{N}(x_t; \sqrt{1-\beta_t} x_{t-1}, \beta_t \mathbf{I})
    \label{eq:singlestep}
\end{alignat}
The sample $x_t$ is obtained by adding \textit{i.i.d.} Gaussian noise with variance $\beta_t$ at timestep $t$ and scaling the previous sample $x_{t-1}$ with $\sqrt{1 - \beta_t}$ according to a variance schedule.

The DDPM is trained to reverse the process in \eqref{eq:singlestep}.
The reverse process is modeled by a neural network that predicts the parameters $\mu_{\theta}(x_t, t)$ and  $\Sigma_{\theta}(x_t, t)$ of a Gaussian distribution,
\begin{equation}
\label{eq:nn}
p_{\theta}(x_{t-1}|x_t) = \mathcal{N}(x_{t-1}; \mu_{\theta}(x_t, t), \Sigma_{\theta}(x_t, t))
\end{equation}
The learning objective for the model \eqref{eq:nn} is derived by considering the variational lower bound,
\begin{align}
\Ea{-\log p_\theta(\bx_0)} \leq \mathbb{E}_{q}\bigg[ &- \log \frac{p_{\theta}(\bx_{0:T})}{q(\bx_{1:T} | \bx_0)} \bigg] \\
= \mathbb{E}_q\bigg[ -\log p(\bx_T) &- \sum_{t \geq 1} \log \frac{p_\theta(\bx_{t-1} | \bx_t)}{q(\bx_t|\bx_{t-1})} \bigg] = L \nonumber
\end{align}
As extended by Ho~\etal~\cite{ddpm}, this loss can be further decomposed as,
\begin{align}
\label{eq:vb}
&\mathbb{E}_q \bigg[ \underbrace{\kl{q(\bx_T|\bx_0)}{p(\bx_T)}}_{L_T} \\
&+ \sum_{t > 1} \underbrace{\kl{q(\bx_{t-1}|\bx_t,\bx_0)}{p_\theta(\bx_{t-1}|\bx_t)}}_{L_{t-1}} \underbrace{-\log p_\theta(\bx_0|\bx_1)}_{L_0} \bigg] \nonumber
\end{align}
Importantly the term $L_{t-1}$ trains the network \eqref{eq:nn} to perform one reverse diffusion step. Furthermore, it allows for a closed from expression of the objective since $q(\bx_{t-1}|\bx_t,\bx_0)$ is also Gaussian~\cite{ddpm}.

As reported by Ho~\etal~\cite{ddpm}, the best way to parametrize the model is to predict the cumulative noise $\epsilon_0$ that is added to the current intermediate image $x_t$. Thus, we obtain the following parametrization of the predicted mean $\mu_\theta(x_t, t)$,
\begin{equation}
\mu_{\theta}(x_t, t) = \frac{1}{\sqrt{\alpha_t}} \left( x_t - \frac{\beta_t}{\sqrt{1-\bar{\alpha}_t}} \epsilon_{\theta}(x_t, t) \right)
\end{equation}
From $L_{t-1}$ in \eqref{eq:vb}, the following simplified training objective is derived by Ho~\etal~\cite{ddpm}, 
\begin{equation}
L_{\text{simple}} = E_{t,x_0,\epsilon}\left[ || \epsilon - \epsilon_{\theta}(x_t, t) ||^2 \right]
\end{equation}

As introduced by Nichol and Dhariwal~\cite{improvedddpm}, learning the variance $\Sigma_{\theta}(x_t, t)$ in \eqref{eq:nn} of the reverse process helps to reduce the number of sampling steps by an order of magnitude. They, therefore, add the variational lower bound loss.
Specifically, we base our training and inference approach on the recent work~\cite{beatGan}, which further reduced the inference time by factor four. 
To train the DDPM, we need a sample $x_t$ and corresponding noise that is used to transform $x_0$ to $x_t$.
By using the independence property of the noise added at each step~\eqref{eq:singlestep}, we can calculate the total noise variance as $\bar{\alpha_t} = \prod_{s=1}^t (1 - \beta_s)$. We can thus rewrite~\eqref{eq:singlestep}, as a single step,
\begin{alignat}{2}
    q(x_t|x_0) &= \mathcal{N}(x_t; \sqrt{\bar{\alpha}_t} x_0, (1-\bar{\alpha}_t) \mathbf{I})
    \label{eq:jumpnoise}
\end{alignat}
It allows us to efficiently sample pairs of training data to train a reverse transition step.

\begin{figure}[t]
   \centering
    
   \includegraphics[width=0.99\linewidth]{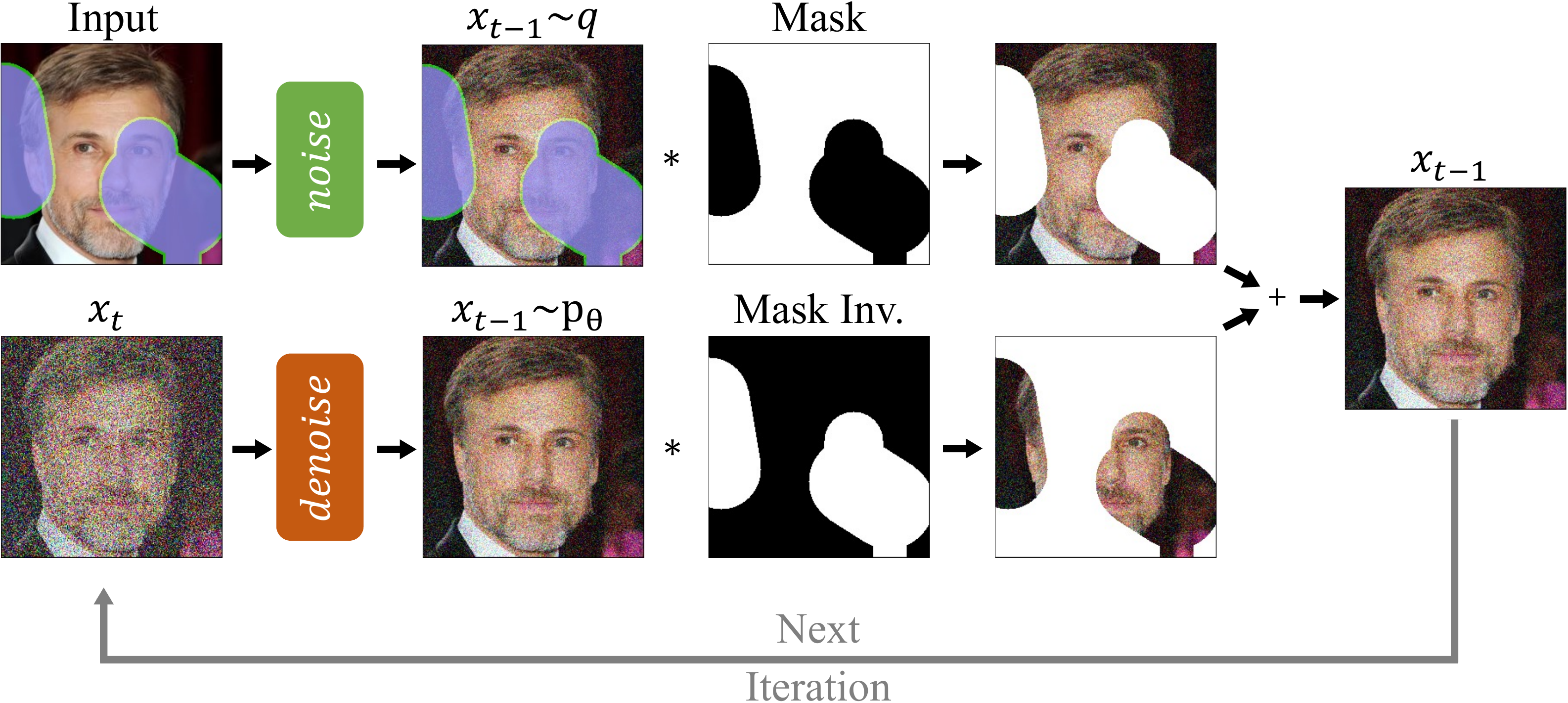}
   
   \caption{\textbf{Overview of our approach.} RePaint modifies the standard denoising process in order to condition on the given image content. In each step, we sample the known region (\textit{top}) from the input and the inpainted part from the DDPM output (\textit{bottom}). }
    \label{fig:method_condition}
    \vspace{-2mm}
\end{figure}

\section{Method}
In this section, we first present our approach for conditioning the reverse diffusion process of an unconditional DDPM for image inpainting in Section~\ref{sec:condition}. Then, we introduce an approach to improve the reverse process itself for inpainting in Section~\ref{sec:resampling}.

\begin{figure*}
    \centering
    \includegraphics[trim=0 0 264px 0,clip,width=\linewidth]{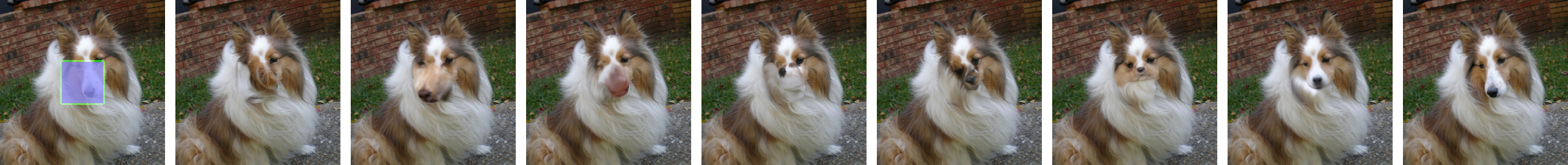}
    \resizebox{\linewidth}{!}{
        \begin{tabular}{ C{2.5cm} C{2.5cm} C{2.5cm} C{2.5cm} C{2.5cm} C{2.5cm} C{2.5cm} C{2.5cm} C{2.5cm} }
           Input & n = 1 & n = 2 & n = 3 & n = 4 & n = 5 & n = 10 & n = 20
    \end{tabular}
    }%
    \caption{\textbf{The effect of applying $n$ sampling steps.} The first example with $n=1$ is the DDPM baseline, the second with $n=2$ is with one resample step. More resampling steps lead to more harmonized images. The benefit saturates at about $n=10$ resamplings.}
    \label{fig:resampling}
\end{figure*}

\subsection{Conditioning on the known Region}
\label{sec:condition}

The goal of inpainting is to predict missing pixels of an image using a mask region as a condition. %
In the remaining of the paper, we consider a trained unconditional denoising diffusion probabilistic model \eqref{eq:nn}. %
We denote the ground truth image as $x$, the unknown pixels as $m \odot x$ and the known pixels as $(1 - m) \odot x$.
\begin{figure}[b]
\vspace{-4mm}
\begin{algorithm}[H]
  \caption{Inpainting using our RePaint approach.} \label{alg:sampling}
  \small
  \begin{algorithmic}[1]
    \vspace{.04in}
    \State $x_T \sim \mathcal{N}(\bzero, \bI)$

    \For{$t=T, \dotsc, 1$}

        \For{$u=1, \dotsc, U$}

          \State $\epsilon \sim \mathcal{N}(\bzero, \bI)$ if $t > 1$, else $\epsilon = \bzero$

          \State $x_{t-1}^\text{known} = \sqrt{\bar{\alpha}_t} x_0 + (1-\bar{\alpha}_t) \epsilon$
          
          \vspace{2mm}

          \State $z \sim \mathcal{N}(\bzero, \bI)$ if $t > 1$, else $\bz = \bzero$

          \State $x_{t-1}^\text{unknown} =
          \frac{1}{\sqrt{\alpha_t}}\left( x_t - \frac{\beta_t}{\sqrt{1-\bar\alpha_t}} \bepsilon_\theta(x_t, t) \right)
          + \sigma_t z$
          
          \vspace{2mm}

          \State $x_{t-1} = m \odot x_{t-1}^\text{known} + (1-m) \odot x_{t-1}^\text{unknown}$
          
          \vspace{2mm}

          \If{$u < U ~\text{and}~ t > 1$}

              \State $x_t \sim \mathcal{N}(\sqrt{1-\beta_{t-1}} x_{t-1}, \beta_{t-1} \mathbf{I})$
              
          \EndIf
          
    \EndFor
    \EndFor
    \State \textbf{return} $x_0$
    \vspace{.04in}
  \end{algorithmic}
\end{algorithm}
\vspace{-0mm}
\end{figure}

Since every reverse step \eqref{eq:nn} from $x_{t}$ to $x_{t-1}$ depends solely on $x_t$, we can alter the known regions $(1 - m) \odot x_{t}$ as long as we keep the correct properties of the corresponding distribution.
Since the forward process is defined by a Markov Chain \eqref{eq:singlestep} of added Gaussian noise, we can sample the intermediate image $x_t$ at any point in time using~\eqref{eq:jumpnoise}. This allows us to sample the know regions $m \odot x_t$ at any time step $t$. Thus, using \eqref{eq:nn} for the unknown region and \eqref{eq:jumpnoise} for the known regions, we achieve the following expression for one reverse step in our approach,
\begin{subequations}
\begin{align}
  x_{t-1}^\text{known} &\sim \mathcal{N}(\sqrt{\bar{\alpha}_t} x_0, (1-\bar{\alpha}_t) \mathbf{I}) \\
  x_{t-1}^\text{unknown} &\sim \mathcal{N}(\mu_{\theta}(x_t, t), \Sigma_{\theta}(x_t, t)) \\
   x_{t-1} &= m \odot x_{t-1}^\text{known} + (1-m) \odot x_{t-1}^\text{unknown}
\end{align} \label{eq:ourStep}
\end{subequations}
Thus, $x_{t-1}^\text{known}$ is sampled using the known pixels in the given image $m \odot x_0$, while $x_{t-1}^\text{unknown}$ is sampled from the model, given the previous iteration $x_t$. These are then combined to the new sample $x_{t-1}$ using the mask. Our approach is illustrated in Figure~\ref{fig:method_condition}.

\begin{figure}
    \centering
    
    \newcommand{\voffCelebA}{-8pt}
    \newcommand{\imgWidthCelebA}{8.25cm}

	\begin{minipage}{10pt}%
    \newcommand{\sizeLeftCelebA}{1.6cm}
	\resizebox{7.9pt}{!}{\rotatebox{90}{
				\begin{tabular}{ C{\sizeLeftCelebA * 2} C{\sizeLeftCelebA * 2} C{\sizeLeftCelebA * 2} C{\sizeLeftCelebA * 2} C{\sizeLeftCelebA * 2} C{\sizeLeftCelebA * 2} }
					Half &
					Expand &
					Alt. Lines &
					~~SR $2\times$ &
					Narrow &
					Wide
				\end{tabular}%
		}}%
	\end{minipage}%
	\begin{minipage}{\imgWidthCelebA-10pt}%
	\includegraphics[width=\textwidth]{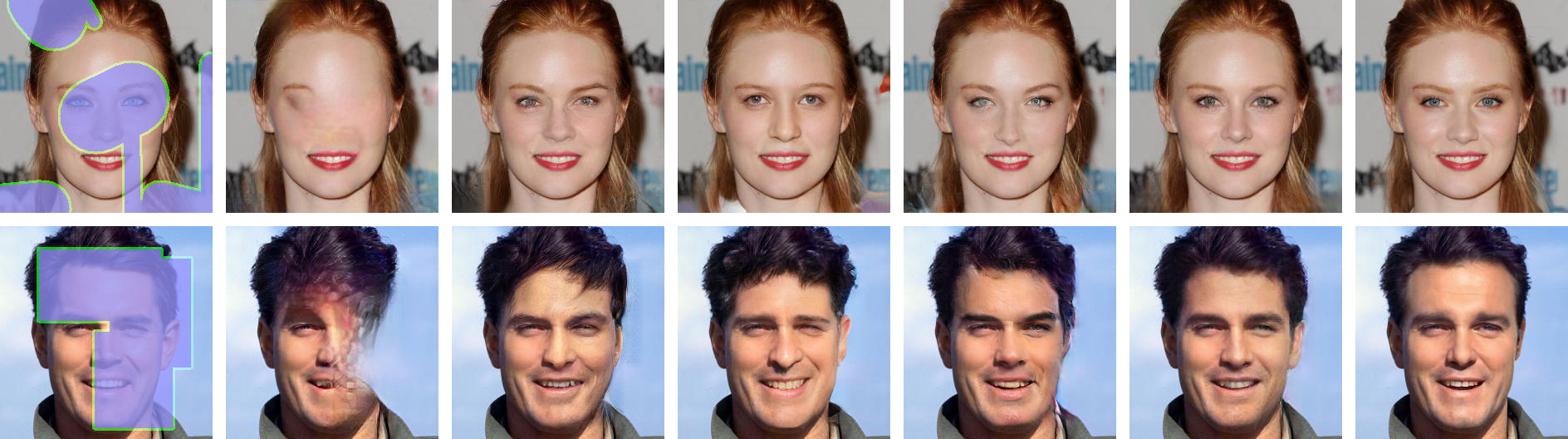} \\
     \vspace{\voffCelebA} \\
	\includegraphics[width=\textwidth]{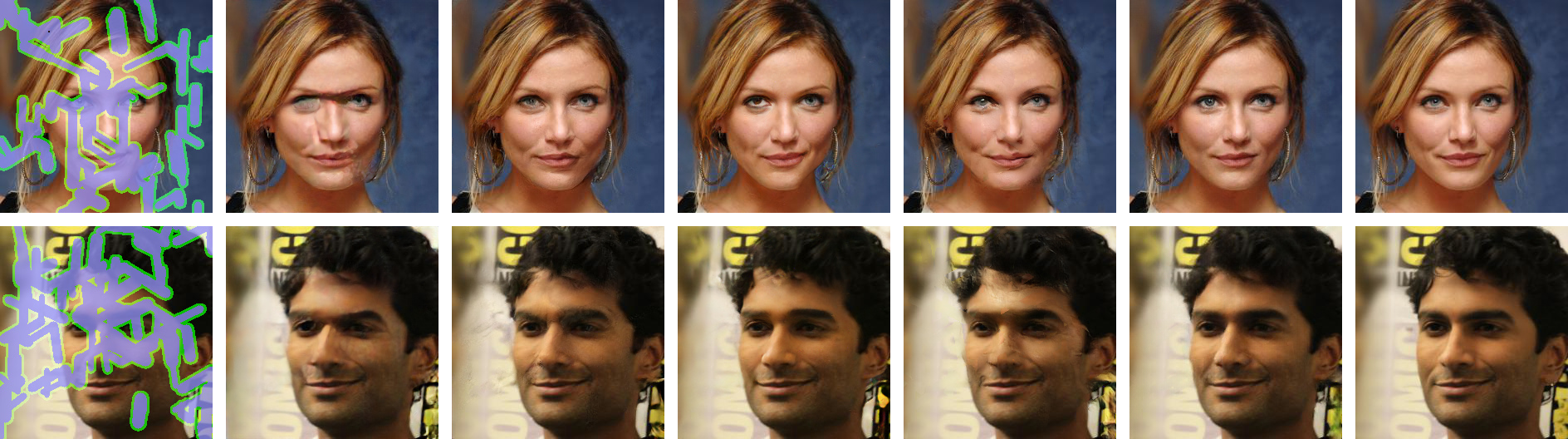} \\
    \vspace{\voffCelebA} \\
	\includegraphics[width=\textwidth]{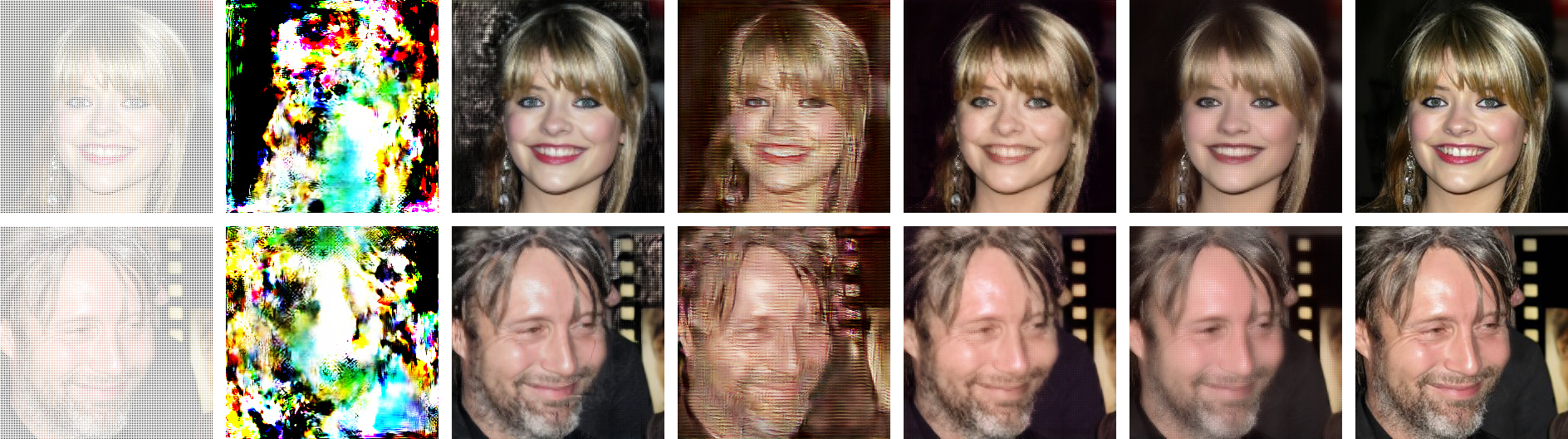} \\
    \vspace{\voffCelebA} \\
    \includegraphics[width=\textwidth]{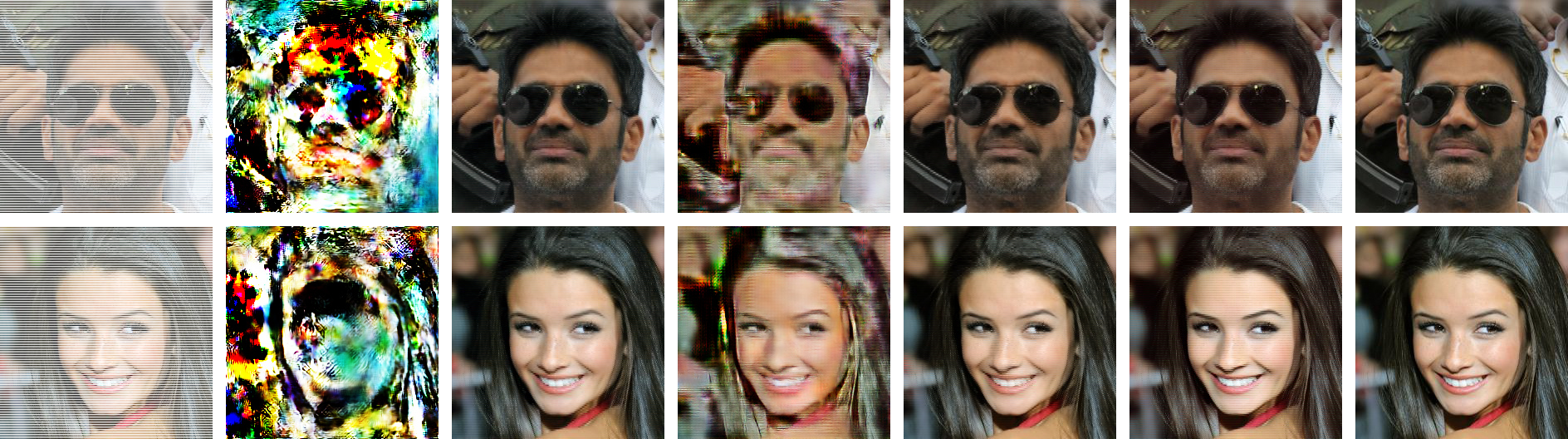} \\
    \vspace{\voffCelebA} \\
    \includegraphics[width=\textwidth]{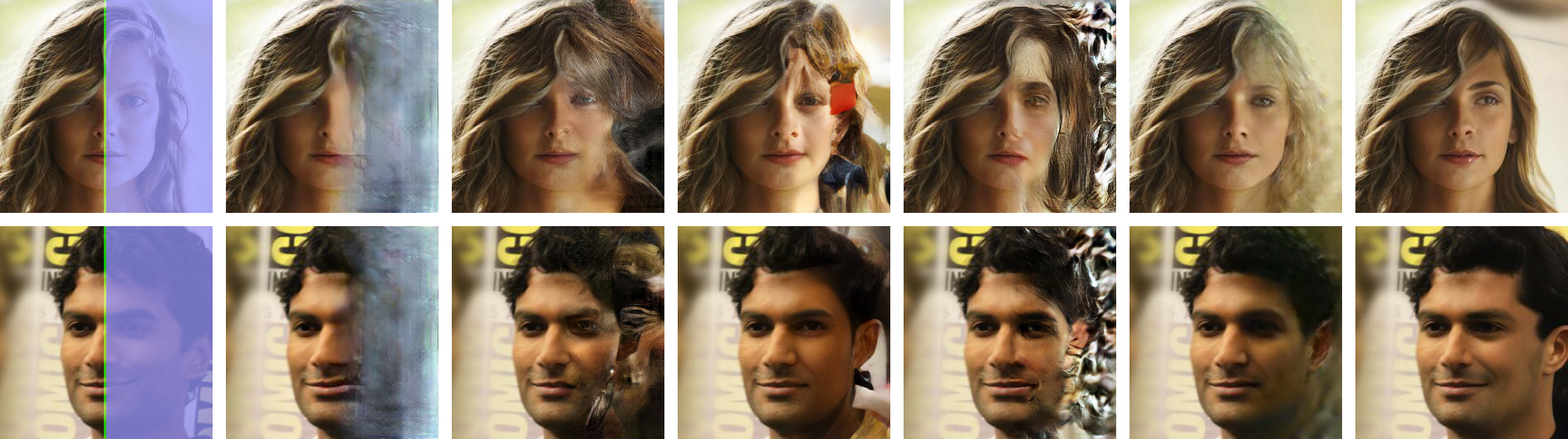} \\
    \vspace{\voffCelebA} \\
    \includegraphics[width=\textwidth]{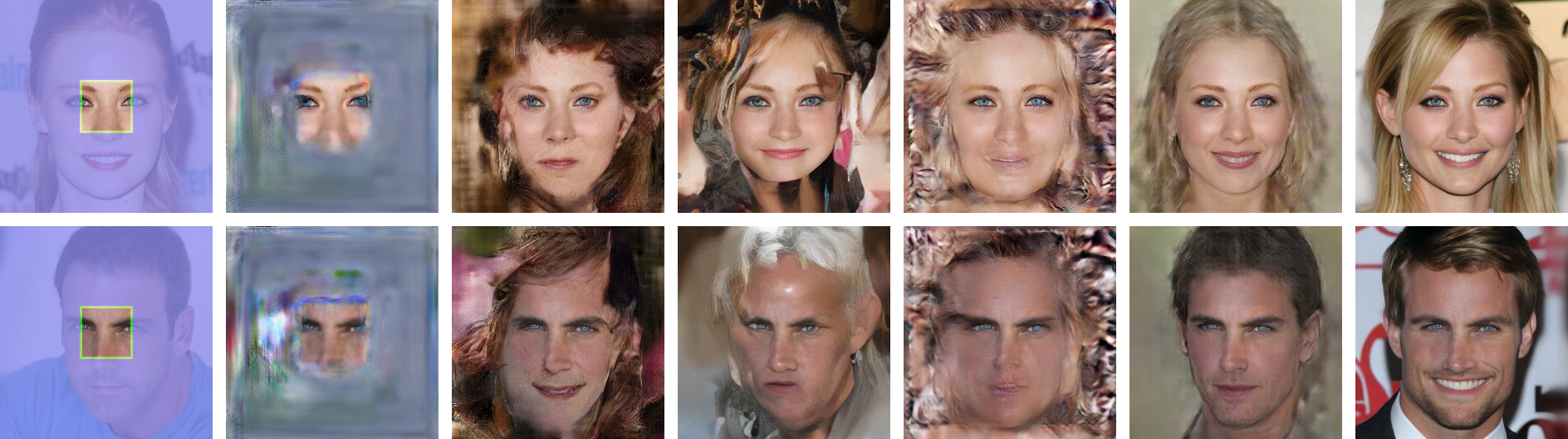}
	\end{minipage}
	\begin{minipage}{10pt}%
	\hspace{\fill}
	\end{minipage}%
	\begin{minipage}{{\imgWidthCelebA-10pt}}%
    \captionRowSeven{1.5}{
    Input &
    AOT~\cite{aot} &
    DSI~\cite{dsi} &
    ICT~\cite{ict} &
    Deep Fill v2~\cite{deepfillv2} &
    LaMa~\cite{lama} &
    \textbf{RePaint} (ours)
    }
	\end{minipage}%
	\vspace{-0mm}
    \caption{
    \textbf{CelebA-HQ Qualitative Results.} Comparison against the state-of-the-art methods for Face Inpainting over several mask settings. Zoom-in for better details.}
    \label{fig:sota_celebA_all}
    \vspace{-4mm}
\end{figure}

\subsection{Resampling}
\label{sec:resampling}

When directly applying the method described in Section~\ref{sec:condition}, we observe that only the content type matches with the known regions.
For example, in Figure~\ref{fig:resampling} $n=1$, the inpainted area is a furry texture matching the hair of the dog. %
Although the inpainted region matches the texture of the neighboring region, it is semantically incorrect. Therefore, the DDPM is leveraging on the context of the known region, yet it is not harmonizing it well with the rest of the image. Next, we discuss possible reasons for this behavior.
From Figure~\ref{fig:method_condition}, we analyze how the method is conditioning the known regions. 
As shown in~\eqref{eq:ourStep}, the model predicts $x_{t-1}$ using $x_t$, which comprises the output of the DDPM~\eqref{eq:nn} and the sample from the known region.
However, the sampling of the known pixels using~\eqref{eq:jumpnoise} is performed without considering the generated parts of the image, which introduces disharmony.
Although the model tries to harmonize the image again in every step, it can never fully converge because the same issue occurs in the next step.
Moreover, in each reverse step, the maximum change to an image declines due to the variance schedule of $\beta_t$.
Thus, the method cannot correct mistakes that lead to disharmonious boundaries in the subsequent steps due to restricted flexibility.
As a consequence, the model needs more time to harmonize the conditional information $x_{t-1}^{\text{known}}$ with the generated information $x_{t-1}^{\text{unknown}}$ in one step before advancing to the next denoising step.

Since the DDPM is trained to generate an image that lies within a data distribution, it naturally aims at producing consistent structures.
In our resampling approach, we use this DDPM property to harmonize the input of the model.
Consequently, we diffuse the output $x_{t-1}$  back to $x_t$ by sampling from \eqref{eq:singlestep} as $x_t \sim \mathcal{N}(\sqrt{1-\beta_t} x_{t-1}, \beta_t \mathbf{I})$.
Although this operation scales back the output and adds noise, some information incorporated in the generated region $x_{t-1}^{\text{unknown}}$ is still preserved in $x_t^{\text{unknown}}$. It leads to a new $x_t^{\text{unknown}}$ which is both more harmonized with $x_t^{\mathtt{known}}$ and contains conditional information from it.

Since this operation can only harmonize one step, it might not be able to incorporate the semantic information over the entire denoising process.
To overcome this problem, we denote the time horizon of this operation as jump length, which is $j=1$ for the previous case.
Similar to the standard change in diffusion speed~\cite{beatGan} (\textit{a.k.a.}~slowing down), the resampling also increases the runtime of the reverse diffusion.
Slowing down applies smaller but more resampling steps by reducing the added variance in each denoising step.
However, that is a fundamentally different approach because slowing down the diffusion still has the problem of not harmonizing the image, as described in our resampling strategy. We empirically demonstrate this advantage of our approach in Sec.~\ref{sec:ablation}.

\section{Experiments}
We perform extensive experiments for face and generic inpainting, compare to the state-of-the-art solutions, and conduct an ablative analysis. In Section~\ref{sec:sota} and~\ref{sec:diversity}, we report a detailed discussion of mask robustness and diversity, respectively. We also report with additional results, analysis, and visuals in the appendix.

\subsection{Implementation Details}
We validate our solution over the CelebA-HQ~\cite{celeba}, and Imagenet~\cite{imagenet} datasets. As our method relies on a pretrained guided diffusion model~\cite{beatGan}, we use the provided ImageNet model. For CelebA-HQ, we follow the same training hyper-parameters as for ImageNet. We use $256 \times 256$ crops in three batches on 4$\times$V100 GPUs each.
In contrast to the pretrained ImageNet model, the CelebA-HQ one is only trained for 250,000 iterations during roughly five days. Note that all our qualitative and quantitative results in the main paper are for 256 image size.

\begin{table*}
    \resizebox{\linewidth}{!}{%
    \begin{tabular}{@{}lrrrrrrrrrrrr@{}}
    \toprule
                           \textbf{CelebA-HQ} & \multicolumn{2}{c}{Wide} & \multicolumn{2}{c}{Narrow} & \multicolumn{2}{c}{Super-Resolve $2\times$} & \multicolumn{2}{c}{Altern. Lines} & \multicolumn{2}{c}{Half} & \multicolumn{2}{c}{Expand} \\
                            Methods &  LPIPS$\downarrow$ &  Votes [\%] &  LPIPS$\downarrow$ &  Votes [\%] &                   LPIPS$\downarrow$ & Votes [\%] &         LPIPS$\downarrow$ &  Votes [\%] &  LPIPS$\downarrow$ &  Votes [\%] &  LPIPS$\downarrow$ &  Votes [\%] \\
    \midrule
                     AOT~\cite{aot} &  0.104 &  $11.6 \pm 2.0$ &  0.047 &  $12.8 \pm 2.1$ &                   0.714 &  $1.1 \pm 0.6$ &         0.667 &   $2.4 \pm 1.0$ &  0.287 &   $9.0 \pm 1.8$ &  0.604 &   $8.3 \pm 1.7$ \\
                     DSI~\cite{dsi} &  0.067 &  $16.0 \pm 2.3$ &  0.038 &  $22.3 \pm 2.6$ &                   0.128 &  $5.5 \pm 1.4$ &         0.049 &   $5.1 \pm 1.4$ &  0.211 &   $4.5 \pm 1.3$ &  0.487 &   $4.7 \pm 1.3$ \\
                     ICT~\cite{ict} &  0.063 &  $27.6 \pm 2.8$ &  0.036 &  $30.9 \pm 2.9$ &                   0.483 &  $4.2 \pm 1.2$ &         0.353 &   $0.7 \pm 0.5$ &  0.166 &  $12.7 \pm 2.1$ &  0.432 &   $8.8 \pm 1.8$ \\
     DeepFillv2~\cite{deepfillv2} &  0.066 &  $23.9 \pm 2.6$ &  0.049 &  $21.0 \pm 2.5$ &                   0.119 &  $9.8 \pm 1.8$ &         0.049 &  $10.6 \pm 1.9$ &  0.209 &   $4.1 \pm 1.2$ &  0.467 &  $13.1 \pm 2.1$ \\
           LaMa~\cite{lama} &  0.045 &  $41.8 \pm 3.1$ &  0.028 &  $33.8 \pm 3.0$ &                   0.177 &  $5.5 \pm 1.4$ &         0.083 &  $20.6 \pm 2.5$ &  0.138 &  $35.6 \pm 3.0$ &  0.342 &  $24.7 \pm 2.7$ \\
                            \textbf{RePaint} &  0.059 &        \textit{Reference} &  0.028 &        \textit{Reference} &                   0.029 &       \textit{Reference} &         0.009 &        \textit{Reference} &  0.165 &        \textit{Reference} &  0.435 &        \textit{Reference} \\
    \bottomrule
    \end{tabular}
    }
        \centering%
        \vspace{0mm}
    \end{table*}
    
\begin{table*}
    \resizebox{\linewidth}{!}{%
    \begin{tabular}{@{}lrrrrrrrrrrrr@{}}
    \toprule
                     \textbf{ImageNet} & \multicolumn{2}{c}{Wide} & \multicolumn{2}{c}{Narrow} & \multicolumn{2}{c}{Super-Resolve $2\times$} & \multicolumn{2}{c}{Altern. Lines} & \multicolumn{2}{c}{Half} & \multicolumn{2}{c}{Expand} \\
                      Methods &  LPIPS$\downarrow$ &  Votes [\%] &  LPIPS$\downarrow$ &  Votes [\%] &                   LPIPS$\downarrow$ &  Votes [\%] &         LPIPS$\downarrow$ &  Votes [\%] &  LPIPS$\downarrow$ &  Votes [\%] &  LPIPS$\downarrow$ &  Votes [\%] \\
    \midrule
               DSI~\cite{dsi} &  0.117 &  $31.7 \pm 2.9$ &  0.072 &  $28.6 \pm 2.8$ &                   0.153 &  $26.9 \pm 2.8$ &         0.069 &  $23.6 \pm 2.6$ &  0.283 &  $31.4 \pm 2.9$ &  0.583 &   $9.2 \pm 1.8$ \\
               ICT~\cite{ict} &  0.107 &  $42.9 \pm 3.1$ &  0.073 &  $33.0 \pm 2.9$ &                   0.708 &   $1.1 \pm 0.6$ &         0.620 &   $6.6 \pm 1.5$ &  0.255 &  $51.5 \pm 3.1$ &  0.544 &  $25.6 \pm 2.7$ \\
     LaMa~\cite{lama} &  0.105 &  $42.4 \pm 3.1$ &  0.061 &  $33.6 \pm 2.9$ &                   0.272 &  $13.0 \pm 2.1$ &         0.121 &   $9.6 \pm 1.8$ &  0.254 &  $41.1 \pm 3.1$ &  0.534 &  $20.3 \pm 2.5$ \\
                      \textbf{RePaint} &  0.134 &        \textit{Reference} &  0.064 &        \textit{Reference} &                   0.183 &        \textit{Reference} &         0.089 &        \textit{Reference} &  0.304 &        \textit{Reference} &  0.629 &        \textit{Reference} \\
    \bottomrule
    \end{tabular}
    }
        \centering%
        \caption{\textbf{CelebA-HQ (\textit{top}) and ImageNet (\textit{bottom}) Quantitative Results.} Comparison against the state-of-the-art methods. We compute the LPIPS (lower is better) and \textit{Votes} for six different mask settings. \textit{Votes} refers to the ratio of votes with respect to ours.}
        \vspace{0mm}%
        \label{tab:sota_inet}
        \vspace{-2mm}
    \end{table*}

For our final approach, we use $T=250$ timesteps, and applied $r=10$ times resampling with jumpy size $j=10$.

\subsection{Metrics}
We compare our RePaint with the baseline methods in a user study described as follows. The user is shown the input image with the blanked missing regions.
Next to this image, we display two different inpainting solutions.
The user is asked to select ``Which image looks more realistic?''.
The user thus evaluates the realism of our RePaint to the result of a baseline.
To avoid biasing the user towards an approach, the methods were anonymized shown in a different random order for each image.
Moreover, each user was asked every question twice and could only submit their answer if they were consistent with themselves in at least 75\% of their answer. A self-consistency in 100\% of the cases is often not possible since, for example, the LaMa method can have a very similar quality to RePaint on the mask settings they provide.
Our user study evaluates all 100 test images of the test datasets CelebA-HQ and ImageNet for the following masks: Wide, Narrow, Every Second Line, Half Image, Expand, and Super-Resolve.
We use the answers of five different humans for every image query, resulting in 1000 votes per method-to-method comparison in each dataset and mask setting, and show the 95\% confidence interval next to the mean votes.
In addition to the user study, we report the commonly reported perceptual metric LPIPS~\cite{lpips}, which is a learned distance metric based on the deep feature space of AlexNet. We compute the LPIPS over the same 100 test images used in the user study. %
The results are shown in Table~\ref{tab:sota_inet}. %
Furthermore, please refer to the appendix for additional quantitative results.%

\begin{figure}[t]
    \centering
    
   \newcommand{\sizeLeftInet}{2.8cm}
   \newcommand{\sizeTextLeft}{6pt}
   \newcommand{\voffInet}{-11pt}

   \begin{minipage}{\sizeTextLeft}%
   \resizebox{\sizeTextLeft}{!}{\rotatebox{90}{
               \begin{tabular}{ C{\sizeLeftInet} C{\sizeLeftInet} C{\sizeLeftInet} C{\sizeLeftInet} C{\sizeLeftInet} C{\sizeLeftInet}} 
   Half & Expand & Altern. Lines & SR $2\times$ & Thin & Thick
               \end{tabular}%
       }}%
   \end{minipage}%
   \begin{minipage}{{\linewidth-\sizeTextLeft}}%
   \includegraphics[width=\linewidth]{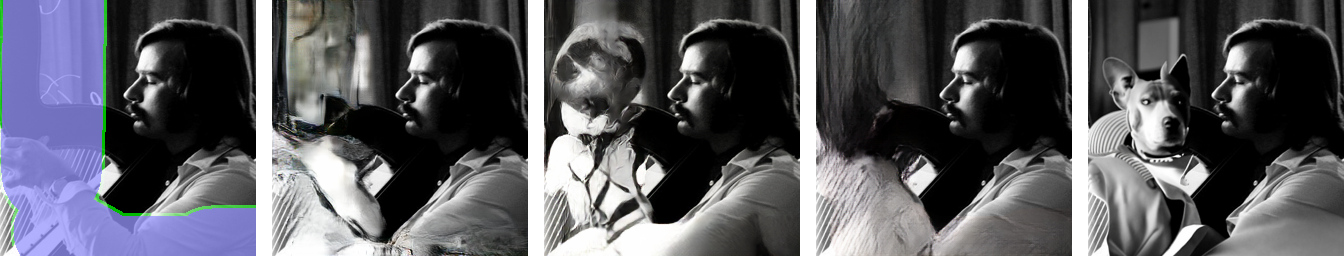} \\
    \vspace{\voffInet} \\
   \includegraphics[width=\linewidth]{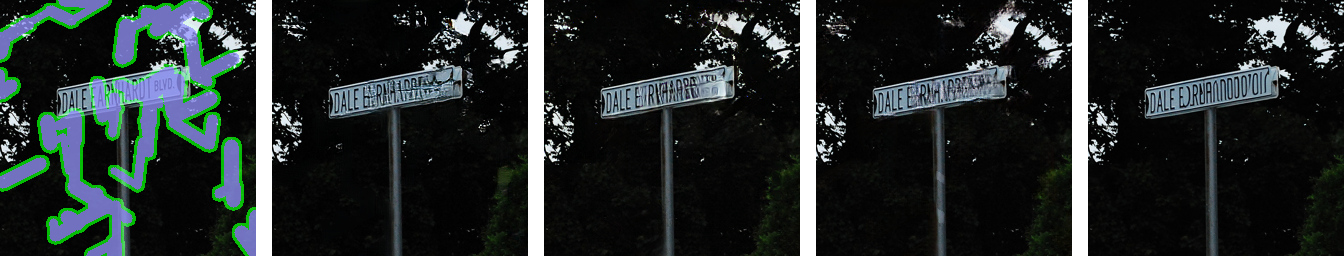} \\
    \vspace{\voffInet} \\
   \includegraphics[width=\textwidth]{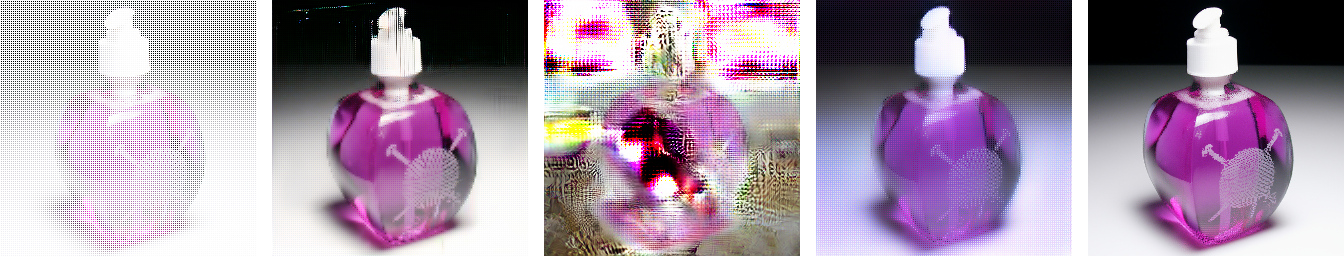} \\
    \vspace{\voffInet} \\
   \includegraphics[width=\textwidth]{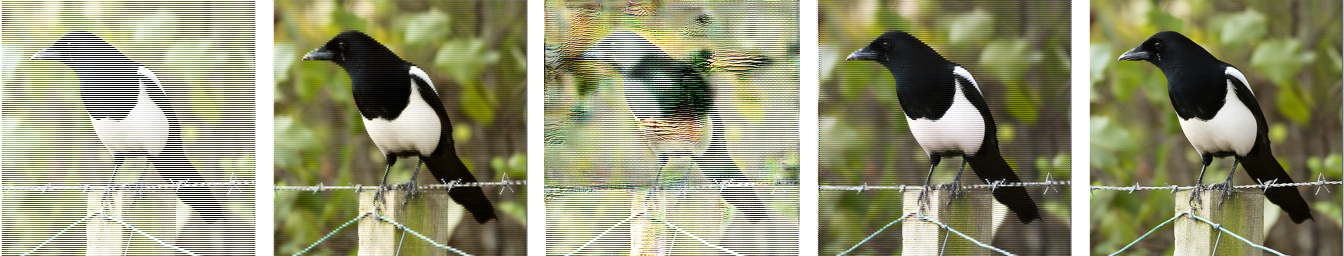} \\
    \vspace{\voffInet} \\
   \includegraphics[width=\textwidth]{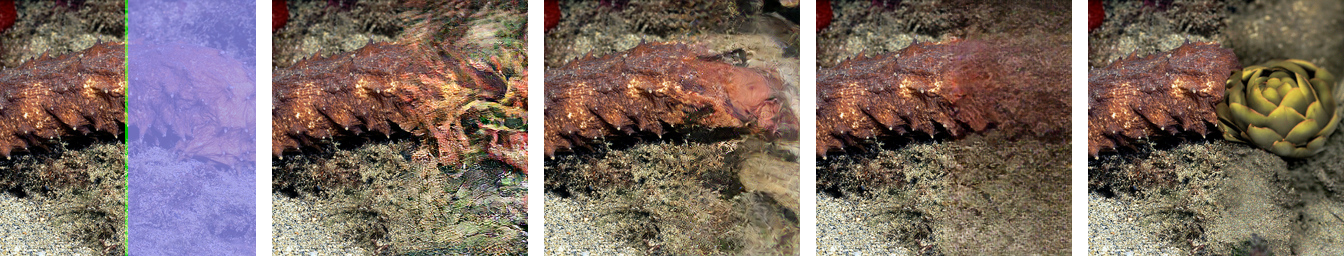} \\
    \vspace{\voffInet} \\
   \includegraphics[width=\textwidth]{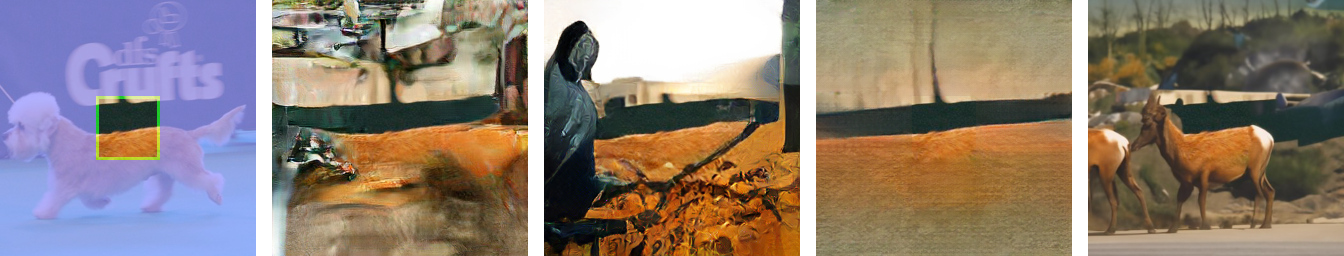}
   \end{minipage}
   \begin{minipage}{\sizeTextLeft}%
   \hspace{\fill}
   \end{minipage}%
   \begin{minipage}{{\linewidth-\sizeTextLeft}}%
   \captionRowSix{2.5}{
   Input & DSI~\cite{dsi} & ICT~\cite{ict} & LaMa~\cite{lama} & \textbf{RePaint} (ours)
   }
   \end{minipage}%
    
   \caption{\textbf{ImageNet Qualitative Results.} Comparison against the state-of-the-art methods for pluralistic inpainting methods over different mask settings. Zoom-in for better details.}
    \label{fig:sota_inet_all}
\vspace{-5mm}
\end{figure}

\subsection{Comparison with State-of-the-Art}
\label{sec:sota}
In this section, we first compare our approach against state-of-the-art on standard mask distributions, commonly employed for benchmarking. We then analyze the generalization capabilities of our method against other approaches. To this end, we evaluate their robustness under four challenging mask settings. Firstly, two different masks that probe if the methods can incorporate information from thin structures. Secondly, two masks that require to inpaint a large connected area of the image. All quantitative results are reported in Table~\ref{tab:sota_inet} and visual results in Figure~\ref{fig:sota_celebA_all} and~\ref{fig:sota_inet_all}.

\parsection{Methods}
We compare our approach against several state-of-the-art autoregressive-based or GAN-based approaches. The autoregressive methods are DSI~\cite{dsi} and ICT~\cite{ict}, and the GAN methods are DeepFillv2~\cite{deepfillv2}, AOT~\cite{aot}, and LaMa~\cite{lama}.
We use their publicly available pretrained models.
We used the existing FFHQ~\cite{karras2019style} pretrained model of ICT for our CelebA-HQ testing. As LaMa does not provide ImageNet models, we trained their system for 300,000 iterations of batch size five using the original implementation. 

\parsection{Settings}
We use 100 images of size 256$\times$256 from CelebA-HQ~\cite{celeba} and ImageNet test sets. The resulting LPIPS and the average votes of the user study are shown in Table~\ref{tab:sota_inet}. %
Additionally, refer to the appendix for qualitative and quantitative results over the Places2~\cite{places2} dataset.

\parsection{Wide and Narrow masks}
To validate our method on the standard image inpainting scenario, we use the LaMa~\cite{lama} settings for Wide and Narrow masks.
RePaint outperforms all other methods with a significance margin of 95\% in both CelebA-HQ and ImageNet, for both Wide and Narrow settings. 
See qualitative results in Figure~\ref{fig:sota_celebA_all} and~\ref{fig:sota_inet_all} and quantitative in Table~\ref{tab:sota_inet}.
The best autoregressive method ICT seems to have less global consistency as observed in Figure~\ref{fig:sota_celebA_all} in the second row, where the eyes do not to match well.
In general, the best GAN approach LaMa~\cite{lama} has better global consistency, yet it produces notorious checkerboard artifacts. Those observations might have influenced the users to vote for RePaint for the majority of images, in which our method generates more realistic images.

\parsection{Thin Masks}
Similar to a Nearest-Neighbor Super Resolution problem, the ``Super-Resolution $2\times$'' mask only leaves pixels with a stride of 2 in height and width dimension, and the ``Alternating Lines'' mask removes the pixels every second row of an image. As seen in Figure~\ref{fig:sota_celebA_all} and~\ref{fig:sota_inet_all}, AOT~\cite{aot} fails completely, while the others either produce blurry images or generate visible artifacts, or both. %
These observations are also confirmed by the user study, where RePaint achieves between 73.1\% and 99.3\% of the user votes.

\parsection{Thick Masks}
The ``Expand'' mask only leaves a center crop of $64 \times 64$ from a $256\times256$ image, and ``Half'' mask, which provides the left half of the image as input. As there is less contextual information, most of the methods struggle (see Figure~\ref{fig:sota_celebA_all} and~\ref{fig:sota_inet_all}). Qualitatively, LaMa comes closer to ours, yet our generated images are sharper and have overall more semantic hallucination.
Noteworthy, LaMa outperforms RePaint in therms of LPIPS on ``Expand'' and ``Half'' for both CelebA and ImageNet (Tab.~\ref{tab:sota_inet}). We argue that this behavior is due to our method being more flexible and diverse in the generation. By generating a semantically different image than that in the Ground-Truth, it makes the LPIPS an unsuitable metric for this particular solution. %

The artifacts produced by the baselines can be explained by strong overfitting to the training masks. In contrast, as our method does not involve mask training, our RePaint can handle any type of mask. In the case of large-area inpainting, RePaint produces a semantically meaningful filling, while others generate artifacts or copy texture.
Finally, RePaint is preferred by the users with confidence 95\% except for the inconclusive result of ICT with ``Half'' masks as shown in Table~\ref{tab:sota_inet}.

\subsection{Analysis of Diversity}
\label{sec:diversity}
As shown in~\eqref{eq:nn}, every reverse diffusion step is inherently stochastic since it incorporates new noise from a Gaussian Distribution. Moreover, as we do not directly guide the inpainted area with any loss, the model is, therefore, free to inpaint anything that semantically aligns with the training set. Figure~\ref{fig:intro} illustrates the diversity and flexibility of our model.

\subsection{Class conditional Experiment}
\begin{figure}
\includegraphics[width=\linewidth]{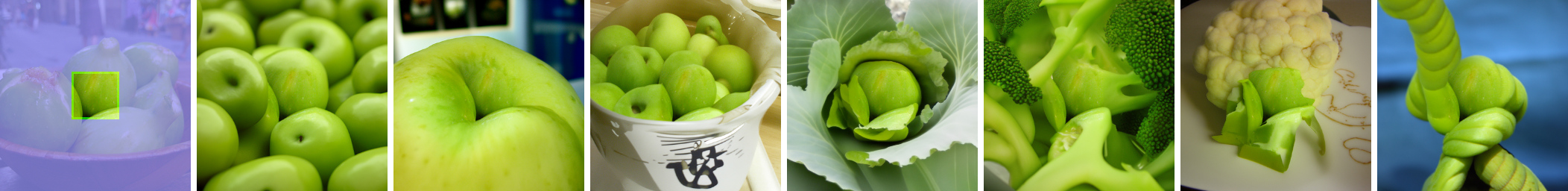}

\vspace{-1.5mm}
\newcommand{\condW}{2.3cm}
\resizebox{\linewidth}{!}{
     \begin{tabular}{ @{}C{\condW}@{}C{6.9cm}@{}C{\condW}@{}C{\condW}@{}C{\condW}@{}C{\condW}@{} }
     Input &
     Apple Samples &
     Head Cabbage &
     Broccoli &
     Cauliflower &
     Knot
\end{tabular}}
\vspace{-5mm}
\captionof{figure}{Visual results for class guided generation on ImageNet.}%
\label{fig:cond}
\vspace{-5mm}
\end{figure}

The pretrained ImageNet DDPM is capable of class-conditional generation sampling. In Figure~\ref{fig:cond} we show examples for the ``Expand'' mask for the ``Granny Smith'' class, as well as other classes. 

\subsection{Ablation Study}
\label{sec:ablation}
\begin{figure}[t]
	\centering
	
   \newcommand{\sizeLeft}{2.1cm}
   \newcommand{\sizeTextLeft}{8pt}

   \begin{minipage}{\sizeTextLeft}%
   \resizebox{\sizeTextLeft}{!}{\rotatebox{90}{
			   \begin{tabular}{ C{\sizeLeft} }
				   Slow-Down
			   \end{tabular}%
	   }}%
   \end{minipage}%
   \begin{minipage}{{\linewidth-\sizeTextLeft}}%
	\includegraphics[width=\linewidth]{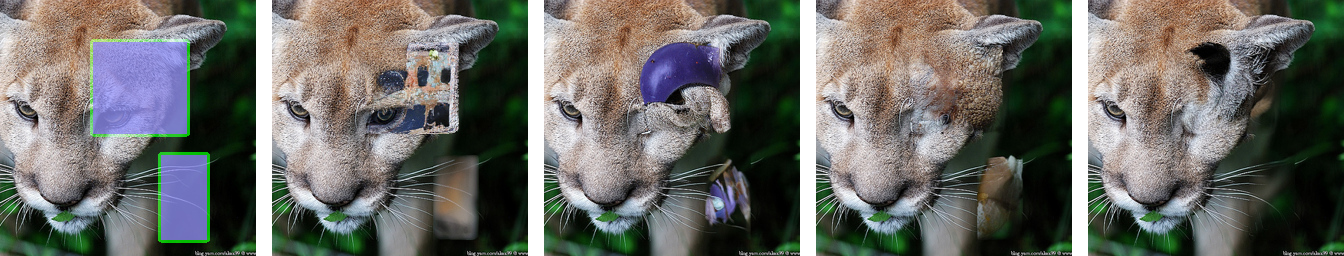}
   \end{minipage}
   \begin{minipage}{\sizeTextLeft}%
   \hspace{\fill}
   \end{minipage}%
   \begin{minipage}{{\linewidth-\sizeTextLeft}}%
   \captionRow{1.8}{
   Input &
   T=250, r=1 &
   T=500, r=1 &
   T=750, r=1 &
   T=1000, r=1
	}%
   \end{minipage}%
   \vspace{0mm}
   \begin{minipage}{\sizeTextLeft}%
   \resizebox{\sizeTextLeft}{!}{\rotatebox{90}{
			   \begin{tabular}{ C{\sizeLeft} C{\sizeLeft} }
				   Resampling
			   \end{tabular}%
	   }}%
   \end{minipage}%
   \begin{minipage}{{\linewidth-\sizeTextLeft}}%
	\includegraphics[width=\linewidth]{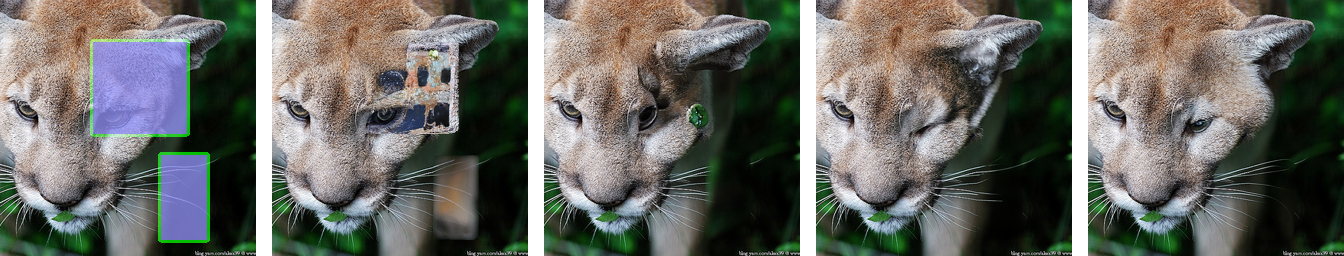}
   \end{minipage}
   \begin{minipage}{\sizeTextLeft}%
   \hspace{\fill}
   \end{minipage}%
   \begin{minipage}{{\linewidth-\sizeTextLeft}}%
   \captionRow{1.8}{
   Input &
   T=250, r=1 &
   T=250, r=2 &
   T=250, r=3 &
   T=250, r=4
	}%
   \end{minipage}%

	\caption{\textbf{Qualitative Analysis of the use of computational budget.} RePaint produces higher visual quality with the same computational budget by resampling (\textit{bottom}) compared to slowing down the diffusion process (\textit{top}). The number of diffusion steps is denoted by $T$ and resamplings by $r$.
	}\vspace{-4mm}%
	\label{fig:ts_vs_resampling}
\end{figure}

\parsection{Comparison to slowing down} To analyze if the increased computational budget causes the improved performance of resampling, we compare it with the commonly used technique of slowing down the diffusion process as described in Section~\ref{sec:resampling}.
Therefore, in Figure~\ref{fig:ts_vs_resampling} and Table~\ref{tab:ts_vs_resampling}, we show a comparison resampling and the slow down in diffusion using the same computational budget for each setting. We observe that the resampling uses the extra computational budget for harmonizing the image, whereas there is no visible improvement at slowing down the diffusion process.

\begin{table}[t]
    \resizebox{\linewidth}{!}{%
    \begin{tabular}{@{}l|rrr|rrr|rrr|rrr@{}}
    \toprule
    & T & r & LPIPS
    & T & r & LPIPS
    & T & r & LPIPS
    & T & r & LPIPS \\
    \midrule
    Slowing down &
    250 & 1 & 0.168 &
    500 & 1 & 0.167 &
    750 & 1 & 0.179 &
    1000 & 1 & 0.161 \\
    Resampling &
    250 & 1 & 0.168 & 
    250 & 2 & 0.148 &
    250 & 3 & 0.142 &
    250 & 4 & 0.134 \\
    \bottomrule
    \end{tabular}
    }
        \centering%
        \caption{
        \textbf{Analysis of the use of computational budget}. We compare slowing down the diffusion process and resampling. We use the ImageNet validation set with 32 images over the LaMa~\cite{lama} Wide mask setting. The number of diffusion steps is denoted by $T$, and the number of resamplings by $r$.}\vspace{-3mm}%
        \label{tab:ts_vs_resampling}
\end{table}

\begin{table}[b]
    \resizebox{\linewidth}{!}{%
    \begin{tabular}{@{}lrrrrrr@{}}
    \toprule
    & \multicolumn{2}{c}{$j = 1$} & \multicolumn{2}{c}{$j=5$}  & \multicolumn{2}{c}{$j=10$} \\
    r   & 
    LPIPS & Votes [\%] &
    LPIPS & Votes [\%] &
    LPIPS & Votes [\%]\\
    \midrule
    5  & 
    0.075 & 42.50$\pm$7.7 &
    0.072 & 46.88$\pm$7.8 &
    0.073 & 53.12$\pm$7.8 \\
    10 & 
    0.088 & 42.50$\pm$7.7 &
    0.073 & 45.62$\pm$7.8 &
    0.068 & 56.25$\pm$7.8 \\
    15 & 
    0.065 & 46.25$\pm$7.8 &
    0.063 & 53.12$\pm$5.5 &
    0.065 & 53.75$\pm$7.8 \\
    \bottomrule
    \end{tabular}
    }
        \centering%
        \caption{
        \textbf{Ablation Study.} Analysis of length of the jumps $j$ and number of resamplings $r$. We report LPIPS and the average user-study votes with respect to LaMa~\cite{lama}. We use 32 images from the CelebA validation set, and use the LaMa Wide mask setting.
        }\vspace{0mm}%
        \label{tab:ablation_jumps}
    \end{table}
    
\parsection{Jumps Length} Moreover, to ablate the jump lengths $j$ and the number of resampling $r$, we study nine different settings in Table~\ref{tab:ablation_jumps}. We obtain better performance at applying the larger jump $j=10$ length than smaller step length steps. We observe that for jump length $j=1$, the DDPM is more likely to output a blurry image. Furthermore, this observation is stable across different numbers of resampling. Furthermore, the number of resamplings increases the performance.

\parsection{Comparison to alternative sampling strategy}
To compare our resampling approach to SDEdit~\cite{sdedit}, we first perform reverse diffusion from $t=T$ to $t=T/2$ to obtain the required initial inpainting at $t=T/2$. We then apply the resampling method from SDEdit, which repeats the reverse process from $t=T/2$ to $t=0$ several times.
The results are shown in Table~\ref{tab:sdedit}.
Our approach achieves significantly better performance across all mask types except for one ``Expand'' case, where LPIPS $>0.6$ is outside a meaningful range for comparisons. In case of `super-resolution masks', our approach reduces the LPIPS by over 53\% on all datasets, clearly demonstrating the advantage of our resampling strategy.

\begin{table}[t]
    \centering\vspace{0mm}%
\resizebox{\linewidth}{!}{%
\begin{tabular}{llrrrrrr}
\toprule
Dataset   & Method         & Wide   & Narrow & Super-Res. & Alt. Lin. & Half   & Expand \\
\midrule
ImageNet  & SDEdit~\cite{sdedit}    & 0.1532 & 0.0952 & 0.3902    & 0.1852    & 0.3272 & \textbf{0.6281} \\
          & RePaint (Ours) & \textbf{0.1341} & \textbf{0.0641} & \textbf{0.1831}    & \textbf{0.0891}    & \textbf{0.3041} & 0.6292 \\
\hline
Places2   & SDEdit~\cite{sdedit}    & 0.1302 & 0.0622 & 0.2712    & 0.1302    & 0.3042 & 0.6202 \\
          & RePaint (Ours) & \textbf{0.1051} & \textbf{0.0441} & \textbf{0.0991}    & \textbf{0.0511}    & \textbf{0.2861} & \textbf{0.6151} \\
\hline
CelebA-HQ & SDEdit~\cite{sdedit}    & 0.0762 & 0.0462 & 0.1132    & 0.0302    & 0.1892 & 0.4492 \\
          & RePaint (Ours) & \textbf{0.0591} & \textbf{0.0281} & \textbf{0.0291}    & \textbf{0.0091}    & \textbf{0.1651} & \textbf{0.4351} \\
\bottomrule
\end{tabular}
}\vspace{-3mm}%
    \caption{Comparison with the resampling schedule proposed in~\cite{sdedit} in terms of LPIPS. The resampling method proposed in our RePaint (Sec.~4.2) achieves substantially better results, in particular for the Super-Resolution masks. }\vspace{0mm}%
    \label{tab:sdedit}\vspace{-5mm}
\end{table}

\section{Limitations}
Our method produces sharp, highly detailed, and semantically meaningful images. We believe that our work opens interesting research directions for addressing the current limitations of the method. Two directions are of particular interest. First, naturally, the per-image DDPM optimization process is significantly slower than the GAN-based and Autoregressive-based counterparts. That makes it currently difficult to apply it for real-time applications. Nonetheless, DDPM is gaining in popularity, and recent publications are working on improving the efficiency~\cite{luhman2021knowledge,luhman2021denoising}. Secondly, for the extreme mask cases, RePaint can produce realistic images completions that are very different from the Ground Truth image. That makes the quantitative evaluation challenging for those conditions. An alternative solution is to employ the FID score~\cite{fid} over a test set. However, a reliable FID for inpainting is usually computed with more than 1,000 images. For current DDPM, this would result in a runtime that is not feasible for most research institutes.

\section{Potential Negative Societal Impact}
On the one hand, RePaint is an inpainting method that relies on an unconditional pretrained DDPM. Therefore, the algorithm might be biased towards the dataset on which it was trained. Since the model aims to generate images of the same distribution as the training set, it might reflect the same biases, such as gender, age, and ethnicity.
On the other hand, RePaint could be used for the anonymization of faces. For example, one could remove the information about the identity of people shown at public events and hallucinate artificial faces for data protection.

\section{Conclusions}
We presented a novel denoising diffusion probabilistic model solution for the image inpainting task. In detail, we developed a mask-agnostic approach that widely increases the degree of freedom of masks for the free-form inpainting. %
Since the novel conditioning approach of RePaint complies with the model assumptions of a DDPM, it produces a photo-realistic image regardless of the type of the mask. %

\parsection{Acknowledgements}
This work was supported by the ETH Z\"urich Fund (OK), a Huawei Technologies Oy (Finland) project, and an Nvidia GPU grant.

\clearpage

{\small
\bibliographystyle{ieee_fullname}
\bibliography{references}
}

\clearpage

\setcounter{section}{0}
\renewcommand{\thesection}{\Alph{section}}

\section*{Appendix}

In this appendix, we provide additional details and analysis of our approach.
We give more explanation on our user study in Section~\ref{sec:user_study}. Further, we present additional details on how we implemented the diffusion time schedule for jumps in Section~\ref{sec:jump}. Visual results for our ablation for jump size and the number of resamplings are provided in Section~\ref{sec:ablation_sup}. 
The evaluation on the second part of the LaMa Benchmark on Places2 is presented in Section~\ref{sec:places2}.
Furthermore, to compare the diversity of the inpaintings for RePaint compared with state-of-the-art, we provide a quantitative analysis in Section~\ref{sec:diversity_sup}. Details on failure cases and data bias on the ImageNet dataset are provided in Section~\ref{sec:failure}. For gaining a better intuitive understanding of the evolution of the latent space, we provide a video of the inference in Section~\ref{sec:video}. And finally, we show additional visual examples in Section~\ref{sec:visual}.

 \begin{figure}[b]
     \centering

    \includegraphics[width=\linewidth]{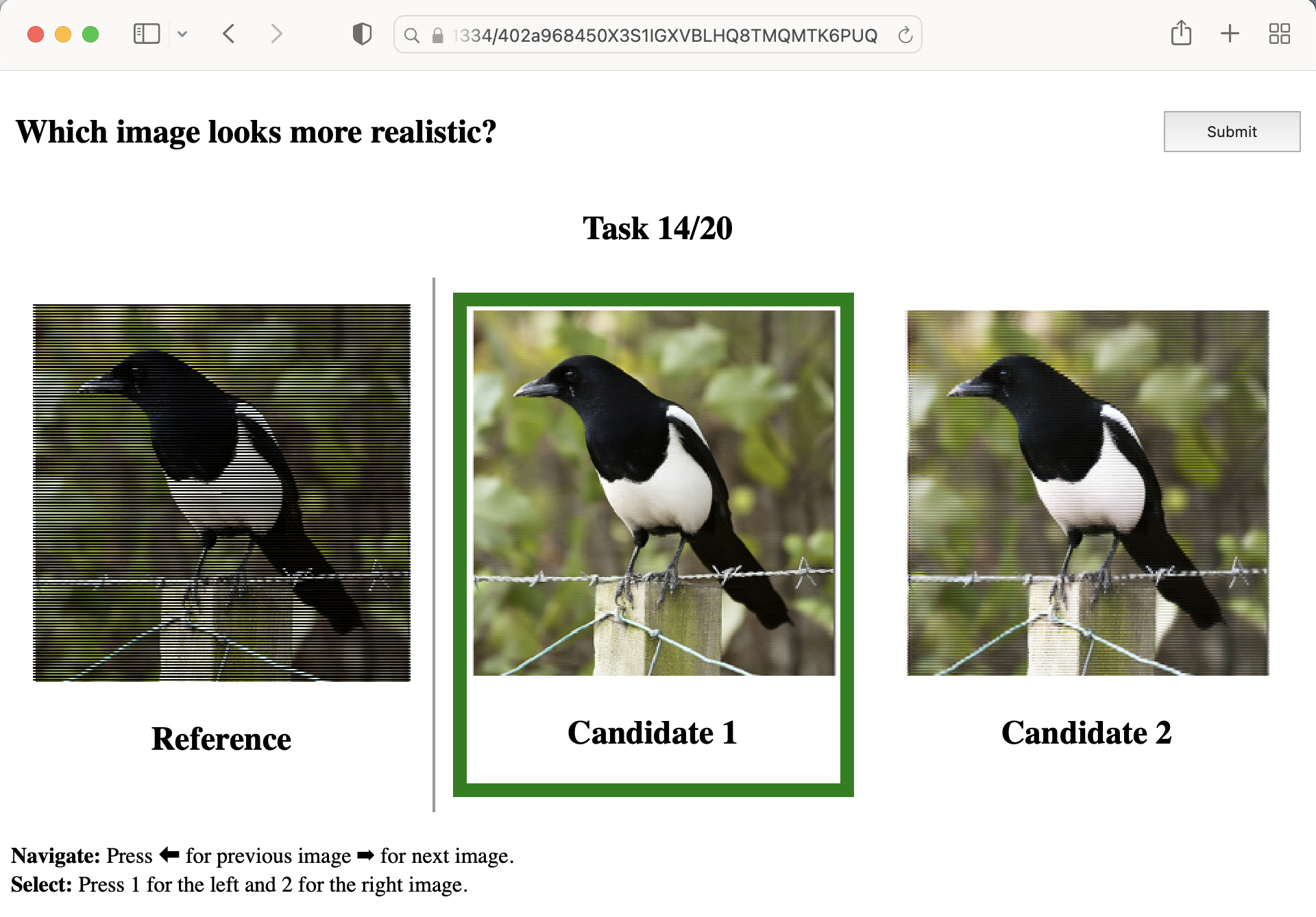}

    \caption{\textbf{User Study Interface.} Example of the user-study interface. Based on the reference image on the Left, the user selects the image that looks more realistic.}
     \label{fig:user_study}
 \end{figure}

\begin{figure}[t]
\begin{python}
t_T = 250
jump_len = 10
jump_n_sample = 10

jumps = {}
for j in range(0, t_T - jump_len, jump_len):
    jumps[j] = jump_n_sample - 1

t = t_T
ts = []

while t >= 1:
    t = t-1
    ts.append(t)

    if jumps.get(t, 0) > 0:
        jumps[t] = jumps[t] - 1
        for _ in range(jump_len):
            t = t + 1
            ts.append(t)

ts.append(-1)
\end{python}
\caption{\textbf{Diffusion Time Schedule.} Pseudo code to generate diffusion time steps for jump length $j=10$ and resample $r=10$.}
\label{fig:algo_ts_generation}
\end{figure}

 \begin{figure*}
     \centering
     
    \includegraphics[width=\linewidth]{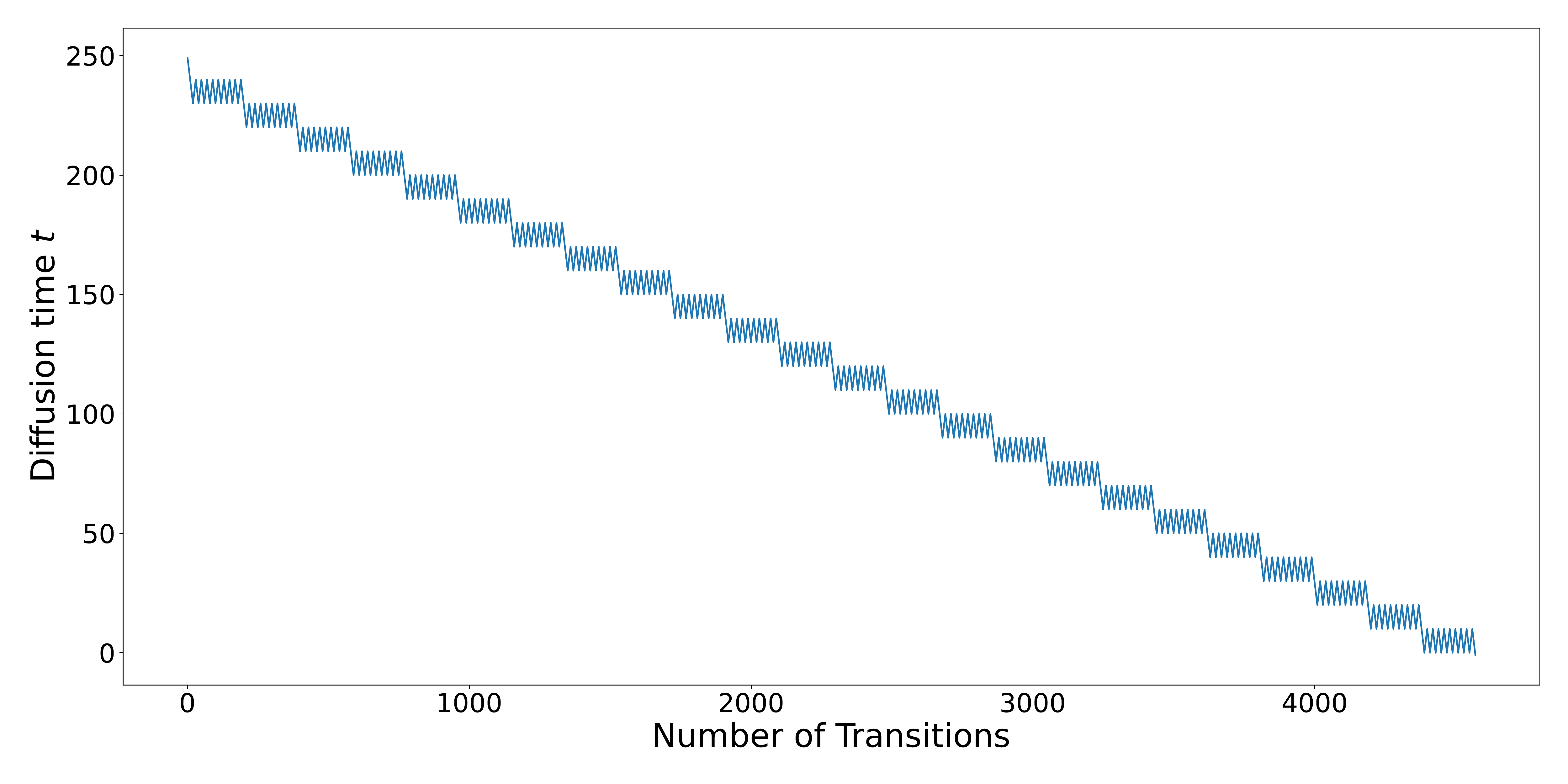}
    
    \caption{\textbf{Diffusion time during inference.} The diffusion time $t$ that a sample $x_t$ is transiting during the inference process with jump length $j=10$ and resampling $r=10$.}
     \label{fig:jump_schedule}
 \end{figure*}

\section{User Study}
\label{sec:user_study}
As described in Section~5.2 in the main paper, we conduct a user study to determine which method is best perceived to the human eye.
In Figure~\ref{fig:user_study}, we depict the user interface, where the user selects the most realistic solution from an input reference. %
To reduce bias, we show the two candidate images in random order. Additionally, to improve the consistency of the user decision and prevent answers with low effort, we show every example twice. The users that agree in less than 75\% of their own votes are discarded. %

\begin{figure}[b]
\begin{python}
times = get_schedule()
x = random_noise()

for t_last, t_cur in zip(times[:-1], times[1:]):
    if t_cur < t_last:
        # Apply Equation 8 (Main Paper)
        x = reverse_diffusion(x, t, x_known)
    else:
        # Apply Equation 1 (Main Paper)
        x = forward_diffusion(x, t)

\end{python}
\caption{\textbf{Inference Process.} Pseudo code of RePaint inference process using a precalculated time schedule.}
\label{fig:algo_jump}
\end{figure}

\section{Algorithm for jump size larger than one}
\label{sec:jump}
In addition to the resampling introduced in Algorithm~1 in the main paper, we use jumps  in diffusion time as described in Section~4.2 in the main paper. Figure~\ref{fig:algo_ts_generation} shows a pseudo-code to further clarify the generation of state transitions. Note that each transition increases or decreases the diffusion time $t$ by one. For example, for a chosen jump length of $j=10$ shown in Figure~\ref{fig:algo_jump}, we apply ten forward transitions before applying ten reverse transitions. The diffusion time $t$ for the latent vector $x_t$ is plotted in Figure~\ref{fig:jump_schedule}.

\begin{table*}
\resizebox{\linewidth}{!}{%
\begin{tabular}{lrrrrrrrrrrrr}
\toprule
                       Datasets & \multicolumn{2}{c}{Wide} & \multicolumn{2}{c}{Narrow} & \multicolumn{2}{c}{Super-Resolve $2\times$} & \multicolumn{2}{c}{Altern. Lines} & \multicolumn{2}{c}{Half} & \multicolumn{2}{c}{Expand} \\
                        Methods &  LPIPS &  Votes [\%] &  LPIPS &  Votes [\%] &                   LPIPS &  Votes [\%] &         LPIPS &  Votes [\%] &  LPIPS &  Votes [\%] &  LPIPS &  Votes [\%] \\
\midrule
                 AOT~\cite{aot} &  0.112 &  $35.4 \pm 3.0$ &  0.062 &  $36.0 \pm 3.0$ &                   0.560 &   $2.2 \pm 0.9$ &         0.399 &   $0.8 \pm 0.6$ &  0.263 &  $34.0 \pm 2.9$ &  0.686 &   $0.7 \pm 0.5$ \\
                 DSI~\cite{dsi} &  0.101 &  $27.4 \pm 2.8$ &  0.054 &  $33.1 \pm 2.9$ &                   0.157 &   $8.4 \pm 1.7$ &         0.083 &   $6.9 \pm 1.6$ &  0.265 &  $33.7 \pm 2.9$ &  0.565 &  $13.8 \pm 2.1$ \\
                 ICT~\cite{ict} &  0.101 &  $35.7 \pm 3.0$ &  0.057 &  $33.7 \pm 2.9$ &                   0.776 &   $0.9 \pm 0.6$ &         0.672 &   $1.3 \pm 0.7$ &  0.256 &  $26.0 \pm 2.7$ &  0.554 &  $26.6 \pm 2.7$ \\
 Deep Fill v2~\cite{deepfillv2} &  0.097 &  $29.7 \pm 2.8$ &  0.051 &  $33.0 \pm 2.9$ &                   0.120 &  $15.8 \pm 2.3$ &         0.070 &  $15.4 \pm 2.2$ &  0.254 &  $32.8 \pm 2.9$ &  0.550 &  $12.9 \pm 2.1$ \\
               LaMa~\cite{lama} &  0.078 &  $47.7 \pm 3.1$ &  0.039 &  $43.3 \pm 3.1$ &                   0.369 &   $7.5 \pm 1.6$ &         0.138 &  $21.5 \pm 2.6$ &  0.233 &  $34.0 \pm 2.9$ &  0.512 &  $39.4 \pm 3.0$ \\
                        RePaint &  0.105 &        \emph{Reference} &  0.044 &        \emph{Reference} &                   0.099 &        \emph{Reference} &         0.051 &        \emph{Reference} &  0.286 &        \emph{Reference} &  0.615 &        \emph{Reference} \\
\bottomrule
\end{tabular}
}
    \centering%
    \caption{\textbf{Places2 Quantitative Results.} We compute the LPIPS (lower is better) and \textit{votes} for five different mask settings. \textit{Votes} refers to the ratio of votes in favor our RePaint.}    
    \label{tab:sota_places2}
    \vspace{0mm}
\end{table*}

\section{Ablation}
\label{sec:ablation_sup}
\begin{figure*}
    \centering
    
    \vspace{-4mm}
    
	\begin{minipage}{0.3\linewidth}
	\includegraphics[width=\textwidth]{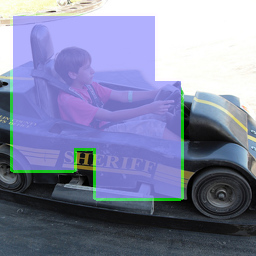}
	
	\vspace{-1mm}
	\centering
	Input
	\end{minipage}%
    
	\begin{minipage}{0.3\linewidth}
	\includegraphics[width=\textwidth]{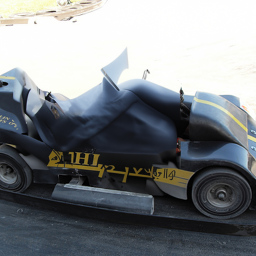}
	
	\vspace{-1mm}
	\centering
	$j=1$~~$r=5$
	\end{minipage}%
	\begin{minipage}{0.02\linewidth}
	\hfill
	\end{minipage}%
	\begin{minipage}{0.3\linewidth}
	\includegraphics[width=\textwidth]{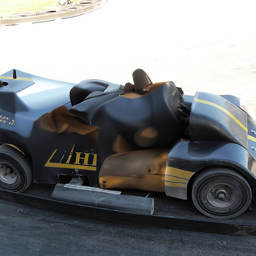}
	
	\vspace{-1mm}
	\centering
	$j=1$~~$r=10$
	\end{minipage}%
	\begin{minipage}{0.02\linewidth}
	\hfill
	\end{minipage}%
	\begin{minipage}{0.3\linewidth}
	\includegraphics[width=\textwidth]{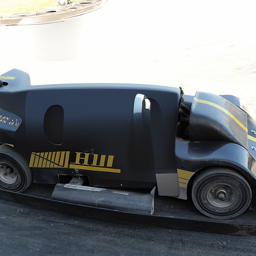}
	
	\vspace{-1mm}
	\centering
	$j=1$~~$r=15$
	\end{minipage}
	
	\begin{minipage}{0.3\linewidth}
	\includegraphics[width=\textwidth]{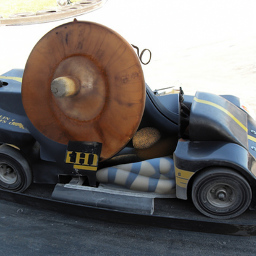}
	
	\vspace{-1mm}
	\centering
	$j=5$~~$r=5$
	\end{minipage}%
	\begin{minipage}{0.02\linewidth}
	\hfill
	\end{minipage}%
	\begin{minipage}{0.3\linewidth}
	\includegraphics[width=\textwidth]{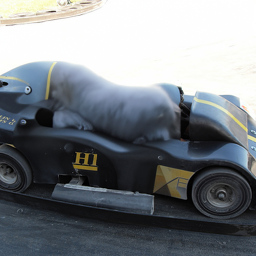}
	
	\vspace{-1mm}
	\centering
	$j=5$~~$r=10$
	\end{minipage}%
	\begin{minipage}{0.02\linewidth}
	\hfill
	\end{minipage}%
	\begin{minipage}{0.3\linewidth}
	\includegraphics[width=\textwidth]{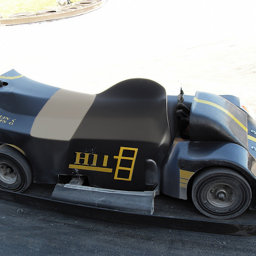}
	
	\vspace{-1mm}
	\centering
	$j=5$~~$r=15$
	\end{minipage}
	
	\begin{minipage}{0.3\linewidth}
	\includegraphics[width=\textwidth]{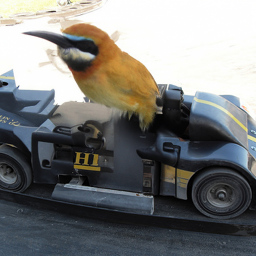}
	
	\vspace{-1mm}
	\centering
	$j=10$~~$r=5$
	\end{minipage}%
	\begin{minipage}{0.02\linewidth}
	\hfill
	\end{minipage}%
	\begin{minipage}{0.3\linewidth}
	\includegraphics[width=\textwidth]{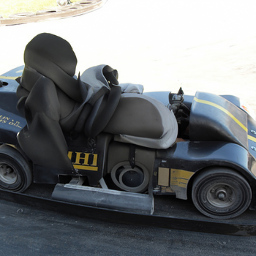}
	
	\vspace{-1mm}
	\centering
	$j=10$~~$r=10$
	\end{minipage}%
	\begin{minipage}{0.02\linewidth}
	\hfill
	\end{minipage}%
	\begin{minipage}{0.3\linewidth}
	\includegraphics[width=\textwidth]{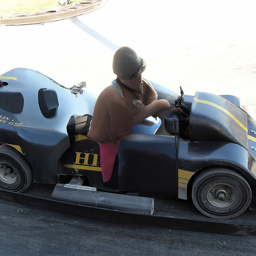}
	
	\vspace{-1mm}
	\centering
	$j=10$~~$r=15$
	\end{minipage}

    \caption{
        \textbf{Ablation Study.} Analysis of length of the jumps $j$ and number of resamplings $r$ on ImageNet validation set with LaMa~\cite{lama} Benchmark mask setting Wide.
        }
    
    \label{fig:ablation_jump}
	
\end{figure*}

In addition to the quantitative analysis in Table~3 in the main paper, this section shows visual examples for different jump lengths $j$ and number or resamplings $r$. As discussed in Section~5.5 in the main paper, smaller jump lengths $j$ tend to produce blurrier images as shown in Figure~\ref{fig:ablation_jump}, and an increased number or resamplings $r$ improves the overall image consistency.

\section{Evaluation on Places2}
\label{sec:places2}
For a more comprehensive experimental framework, in this section, we provide the second part of the benchmark proposed in LaMa~\cite{lama}, which is over the Places2~\cite{places2} dataset.
The experiments on Places2 were conducted using an unconditional model that we trained for 300k iterations with batch size four on four V100, taking about six days in total. 
All other training settings were kept as originally~\cite{beatGan} used for ImageNet.
The model checkpoint will be published. We will clarify these aspects and add further details in the paper.
We use the same mask generation procedure and settings described in the main paper. The results shown in Table~\ref{tab:sota_places2} are in line with those on CelebA and ImageNet in Table~1 of the main paper. RePaint outperforms all other methods for all masks with significance 95\% except for one inconclusive case. 
This case is when comparing RePaint to LaMa on Wide Masks, where the users vote in 52.4\% for RePaint, but the significance interval overlaps with the 50\% border.
The visual comparison on the and Wide and Narrow mask is shown in Figure~\ref{fig:sota_places2_lama}. Moreover, the visual results further confirm the robustness against sparse masks as shown in Figure~\ref{fig:sota_places2_sparse}. The mask pattern is clearly visible in all competing methods, while RePaint shows better harmonization. Regarding large masks, RePaint is able to inpaint semantically meaningful content such as the companion in the Bar in the same age, and overall lightning conditions as shown in the second row of Figure~\ref{fig:sota_places2_huge}.

\begin{table}
\resizebox{\linewidth}{!}{%
\begin{tabular}{lrrrrrrrrrrrr}
\toprule
Mask & \multicolumn{2}{c}{Wide} & \multicolumn{2}{c}{Narrow} & \multicolumn{2}{c}{SR 2x} & \multicolumn{2}{c}{Alter. Lines} & \multicolumn{2}{c}{Half} & \multicolumn{2}{c}{Expand} \\
Measure &   LPIPS &  DS &   LPIPS &  DS &   LPIPS &  DS &        LPIPS &  DS &   LPIPS &  DS &   LPIPS &  DS \\
\midrule
DSI\cite{dsi}     &  0.0639 &  16.68 &  0.0454 &  18.74 &  0.1404 &  12.38 &       0.0591 &   4.78 &  0.2348 &  15.30 &  0.5458 &  14.33 \\
ICT\cite{ict}     &  0.0596 &  15.77 &  0.0402 &  18.65 &  0.5427 &   8.70 &       0.3916 &   8.16 &  0.1817 &  16.40 &  0.4779 &  17.25 \\
RePaint &  0.0552 &  16.40 &  0.0337 &  23.79 &  0.0327 &  19.84 &       0.0106 &  23.00 &  0.1839 &  17.31 &  0.4832 &  17.11 \\
\bottomrule
\end{tabular}
}
    \centering%
    \caption{\textbf{Diversity Score.} The Diversity Score (DS) and LPIPS calculated on CelebA-HQ on various masks for 32 images.}    
    \label{tab:diversity}
    \vspace{-4mm}
\end{table}

\begin{figure*}
    \centering
    
    \newcommand{\sizeLeft}{2cm}

	\begin{minipage}{10pt}%
	\resizebox{10pt}{!}{\rotatebox{90}{
				\begin{tabular}{ C{\sizeLeft} C{\sizeLeft} C{\sizeLeft} C{\sizeLeft} }
					Half &
					Expand
				\end{tabular}%
		}}%
	\end{minipage}%
	\begin{minipage}{{\textwidth-11pt}}%
    \includegraphics[width=\textwidth]{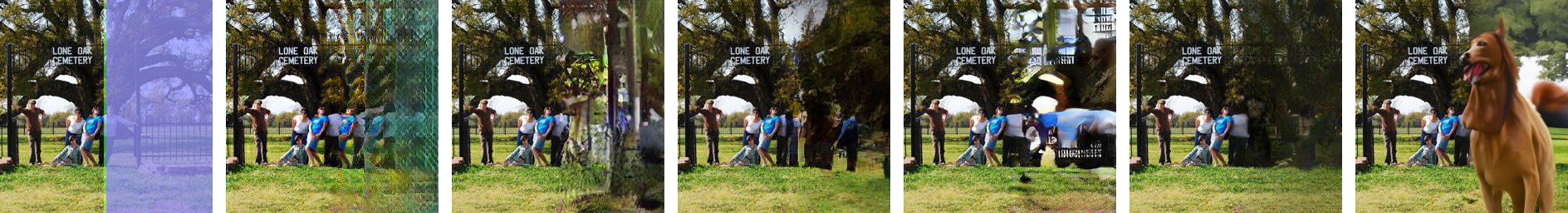}
    \includegraphics[width=\textwidth]{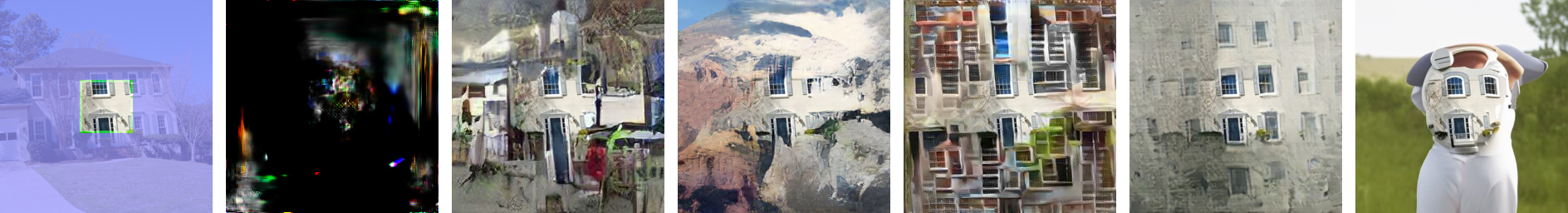}
	\end{minipage}
	\begin{minipage}{10pt}%
	\hspace{\fill}
	\end{minipage}%
	\begin{minipage}{{\textwidth-10pt}}%
    \captionRowEight{2.5}{
    Input &
    AOT~\cite{aot} &
    DSI~\cite{dsi} &
    ICT~\cite{ict} &
    Deep Fill v2~\cite{deepfillv2} &
    LaMa~\cite{lama} &
    \textbf{RePaint} (ours)
    }
	\end{minipage}%
	
    \caption{\textbf{Failure Cases on ImageNet.} When applying RePaint trained on ImageNet for inpainting it is more likely to inpaint dogs, due to the data bias. Zoom-in for better details.
    }

    \label{fig:failure}
\end{figure*}

\section{Diversity}
\label{sec:diversity_sup}
For our quantitative evaluation in the main paper, we sample a single image per input. However, since our method is stochastic, we can sample from it.
To compare the diversity among the stochastic methods, we use the Diversity Score as described in~\cite{ntire21srspace} (higher is better). In contrast to the standard diversity metric~\cite{dsi,ict} that only computes the mean LPIPS across pair of outputs, 
this score is designed to describe meaningful diversity yet also weighting the overall performance in LPIPS.
It aims at measuring the diversity of the generations inside the manifold of plausible predictions. In detail, too extreme predictions or failures are therefore penalized.
As shown in Table~\ref{tab:diversity}, for ``Wide'' and ``Half'', there is no method with both best LPIPS and Diversity Score and for ``Expand'' ICT beats RePaint by $0.81\%$ in Diversity Score and $1.1\%$ in LPIPS. RePaint is superior by a large margin in both LPIPS and Diversity Score for the thin structured masks ``Narrow'', ``Super-Resolution $2\times$'', and ``Alternating Lines'' to both ICT~\cite{ict} and DSI~\cite{dsi}.

\begin{figure}[b]
    \centering
    
   \href{https://www.dropbox.com/s/u64lpnmrnp3a4g2/RePaint_Diffusion_Process.mp4?dl=1}{
   \includegraphics[width=\linewidth]{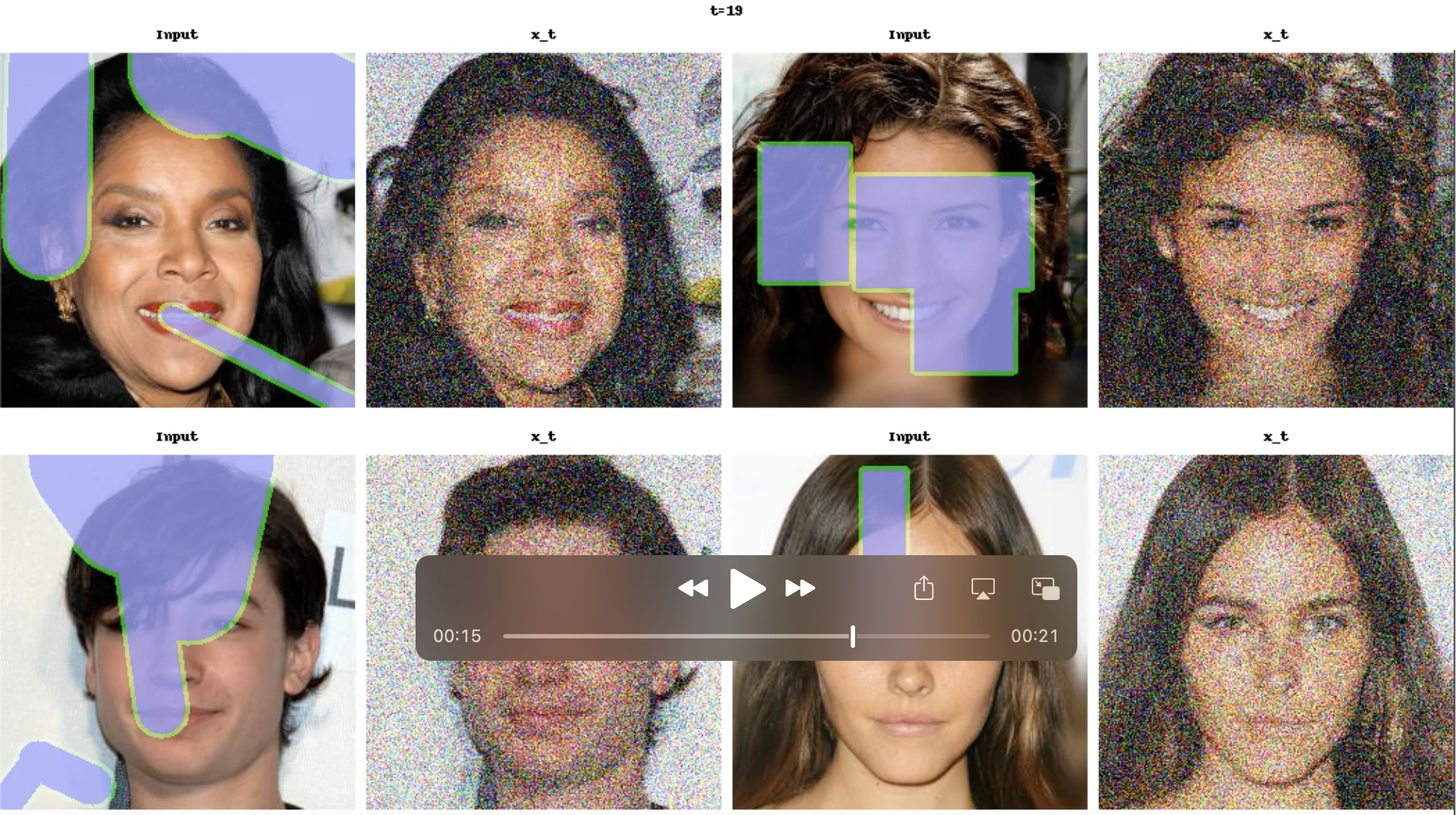}
    }
   
   \caption{\textbf{Video of Diffusion Process.} In the attachment we show the video of the denoising diffusion process on the CelebA-HQ validation set.}    
    \label{fig:video}
\end{figure}

\section{Failure Cases}
\label{sec:failure}
As depicted in Figure~\ref{fig:failure}, RePaint sometimes confuses the semantic context and mixes non-matching objects.
Our model on ImageNet seems to be biased towards inpainting dogs more frequently than expected. 
Since ImageNet has many different breeds of dogs for classification tasks, dogs are over-represented in the training set, hence our model bias. %

\section{Attached Video}
\label{sec:video}
To inspect the latent space of the diffusion space, we provide a video in the attachment as shown in the screenshot in Figure~\ref{fig:video}.
There we show the Ground Truth and the latent space $x_t$ after every transition in the diffusion process. Note that the diffusion time $t$, shown on top, jumps up and down according to the following schedule:
The jump length is $j=5$, and the number of resamplings is $r=9$. To focus more on the visually interesting part of the diffusion process we set the number of diffusion steps to $T=100$ and start resampling below $t=50$.

\begin{figure}
\includegraphics[width=\linewidth]{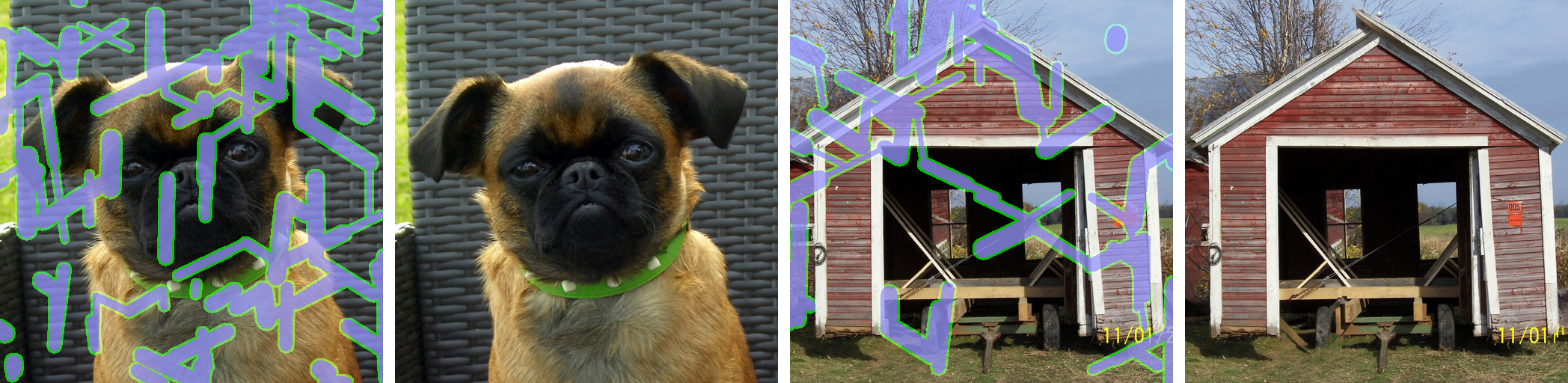}

\vspace{-1mm}
 \resizebox{\linewidth}{!}{
     \newcommand{\inetW}{3.5cm}
     \begin{tabular}{ @{}C{\inetW}@{}C{\inetW}@{}C{\inetW}@{}C{\inetW}@{} }
     Input &
     RePaint (Ours) & 
     Input &
     RePaint (Ours)
\end{tabular}}
\vspace{-2mm}
\captionof{figure}{Visual results on ImageNet $512 \times 512$ for thin mask.}%
\label{fig:inet512}
\vspace{-1mm}
\end{figure}

\section{Experiment on larger resolution}
As shown in Figure~\ref{fig:inet512}, our inpainting method also works on pretrained model from~\cite{beatGan} for $512\times512$. However, we were not able to conduct our full analysis on that resolution due to limited computational resources.

\section{Additional Visual Results}
\label{sec:visual}
We also provide additional visual examples for CelebA-HQ and ImageNet, comparing our approach to the same state-of-the-art methods as in the main paper. We show the results for Wide and Narrow masks in Figures~\ref{fig:sota_celebA_lama} and~\ref{fig:sota_inet_lama}, respectively, for the sparse masks  ``Super-Resolve $2\times$'' and ``Alternating Lines'' in Figures~\ref{fig:sota_celebA_sparse} and~\ref{fig:sota_inet_sparse} and for ``Half'' and ``Expand'' in Figures~\ref{fig:sota_celebA_huge} and~\ref{fig:sota_inet_huge}.

\def \dsname {CelebA-HQ~}
\def \dstag {c}

\begin{figure*}
    \centering
    
    \newcommand{\voffCelebA}{-8pt}
    \newcommand{\imgWidthCelebA}{15.7cm}
    
    \vspace{-5mm}
	\begin{minipage}{10pt}%
    \newcommand{\sizeLeftCelebA}{6.4cm}
	\resizebox{10pt}{!}{\rotatebox{90}{
				\begin{tabular}{ C{\sizeLeftCelebA * 2} C{\sizeLeftCelebA * 2} C{\sizeLeftCelebA * 2} C{\sizeLeftCelebA} C{\sizeLeftCelebA} C{\sizeLeftCelebA} }
					Narrow &
					Wide
				\end{tabular}%
		}}%
	\end{minipage}%
	\begin{minipage}{\imgWidthCelebA-10pt}%
	\includegraphics[width=\textwidth]{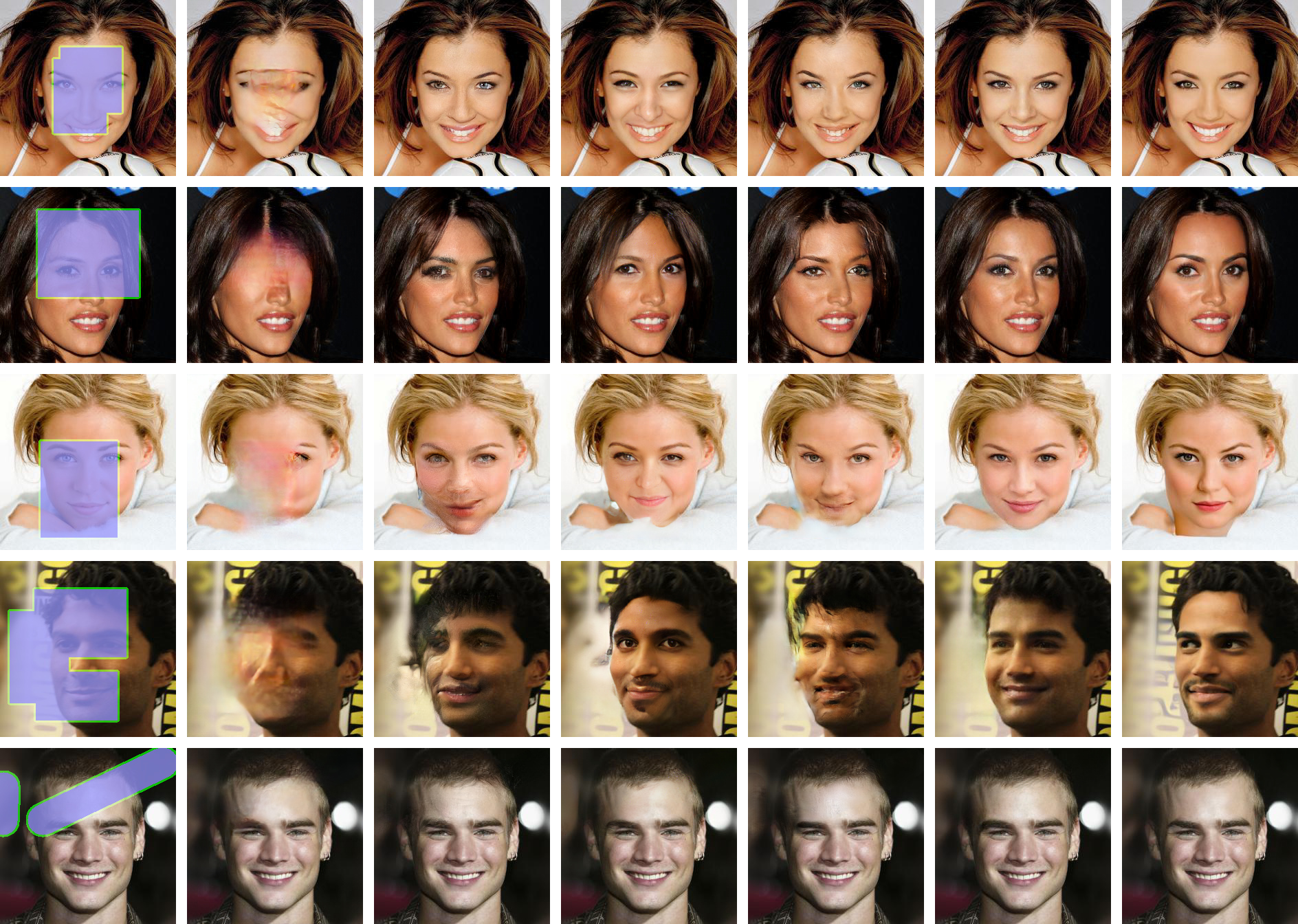} \\
    \vspace{\voffCelebA} \\
	\includegraphics[width=\textwidth]{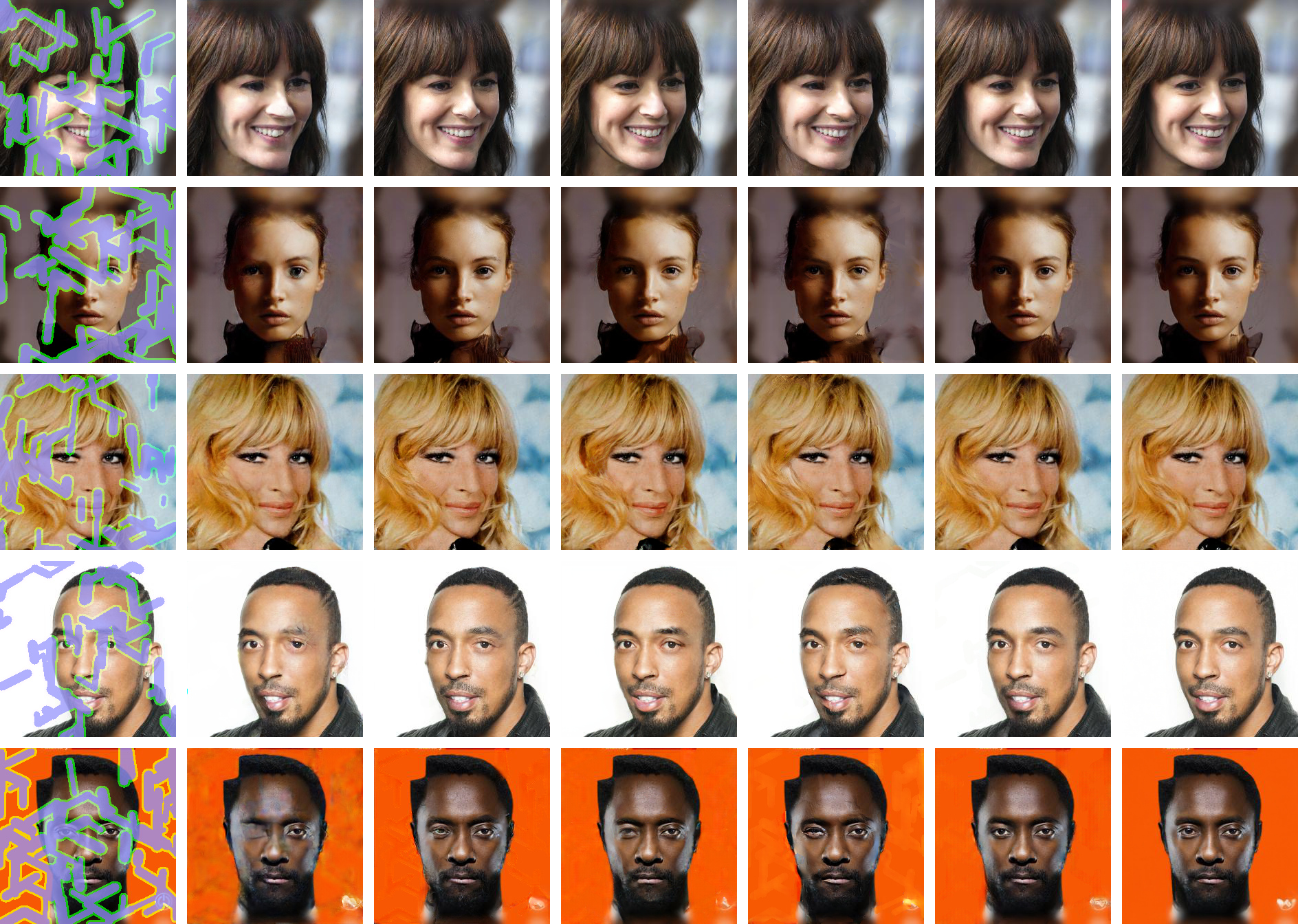} \\
	\end{minipage}
	\begin{minipage}{10pt}%
	\hspace{\fill}
	\end{minipage}%
	\begin{minipage}{{\imgWidthCelebA-10pt}}%
    \captionRowEight{2.5}{
    Input &
    AOT~\cite{aot} &
    DSI~\cite{dsi} &
    ICT~\cite{ict} &
    Deep Fill v2~\cite{deepfillv2} &
    LaMa~\cite{lama} &
    \textbf{RePaint} (ours)
    }
	\end{minipage}%
	\vspace{-0mm}
    \caption{
    \textbf{\dsname Qualitative Results.} Comparison against the state-of-the-art methods for face inpainting. Zoom for better details.}%
    \label{fig:sota_celebA_lama}
\end{figure*}

\begin{figure*}
    \centering
    
    \newcommand{\voffCelebA}{-8pt}
    \newcommand{\imgWidthCelebA}{15.7cm}
    
    \vspace{-5mm}
	\begin{minipage}{10pt}%
    \newcommand{\sizeLeftCelebA}{5.2cm}
	\resizebox{10pt}{!}{\rotatebox{90}{
				\begin{tabular}{ C{\sizeLeftCelebA * 2} C{\sizeLeftCelebA * 2} C{\sizeLeftCelebA * 2} C{\sizeLeftCelebA} C{\sizeLeftCelebA} C{\sizeLeftCelebA} }
					Alternating Lines &
					Super-Resolution $2\times$
				\end{tabular}%
		}}%
	\end{minipage}%
	\begin{minipage}{\imgWidthCelebA-10pt}%
	\includegraphics[width=\textwidth]{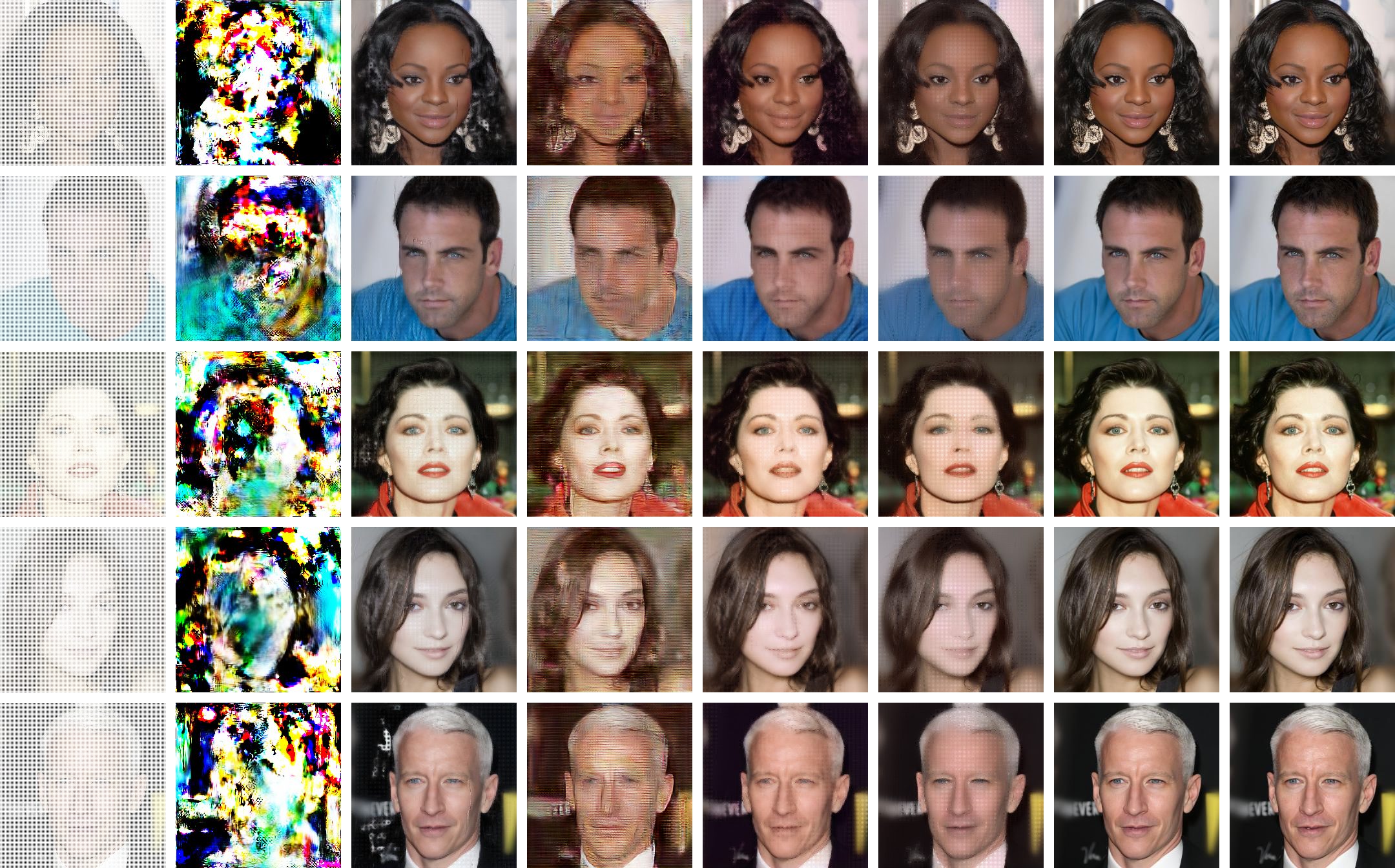} \\
    \vspace{\voffCelebA} \\
	\includegraphics[width=\textwidth]{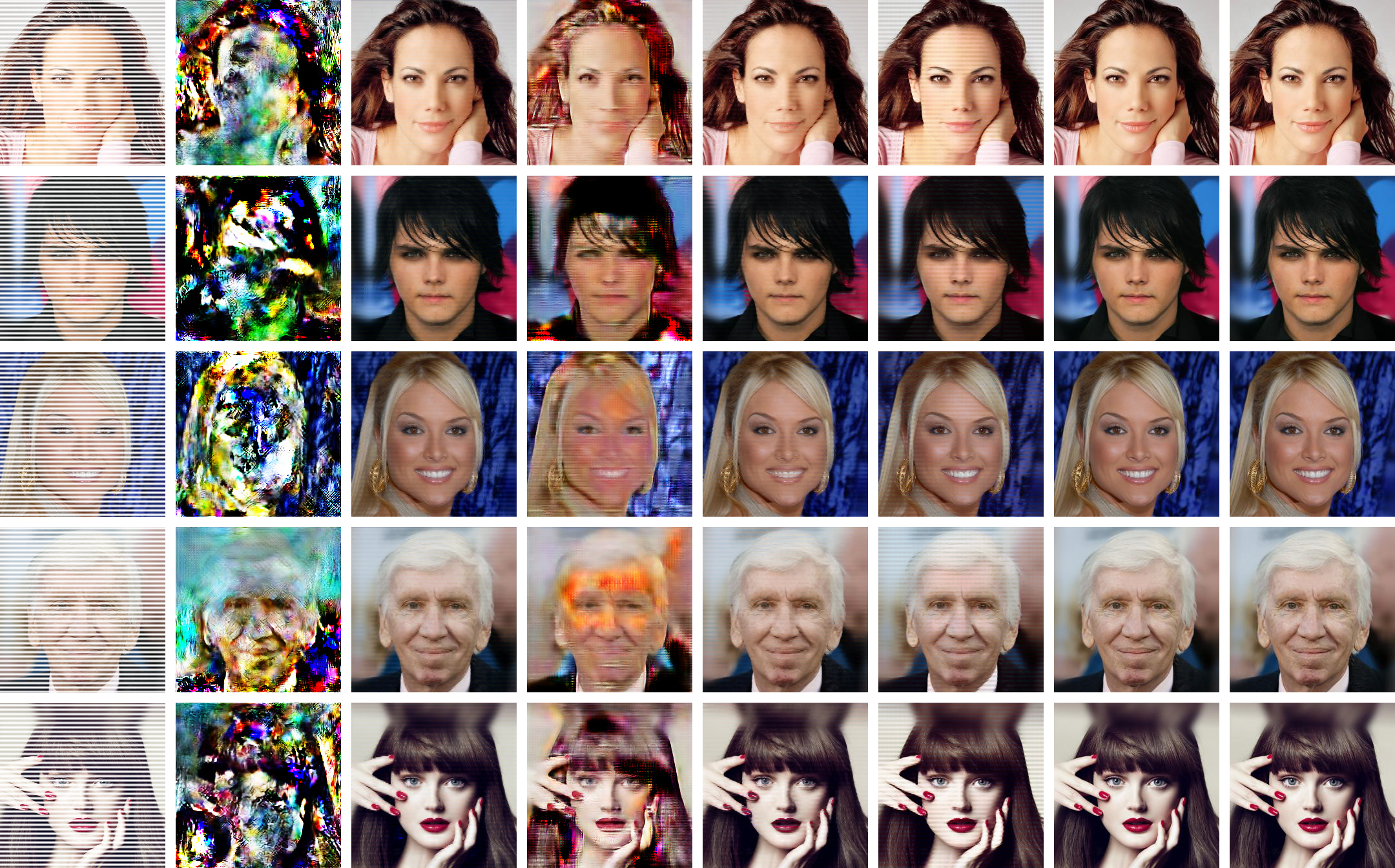} \\
	\end{minipage}
	\begin{minipage}{10pt}%
	\hspace{\fill}
	\end{minipage}%
	\begin{minipage}{{\imgWidthCelebA-10pt}}%
    \captionRowEight{2.5}{
    Input &
    AOT~\cite{aot} &
    DSI~\cite{dsi} &
    ICT~\cite{ict} &
    Deep Fill v2~\cite{deepfillv2} &
    LaMa~\cite{lama} &
    \textbf{RePaint} (ours) &
    Ground Truth
    }
	\end{minipage}%
	\vspace{-0mm}
    \caption{
    \textbf{\dsname Qualitative Results.} Comparison against the state-of-the-art methods for face inpainting. Zoom for better details.}%
    \label{fig:sota_celebA_sparse}
\end{figure*}

\begin{figure*}
    \centering
    
    \newcommand{\voffCelebA}{-8pt}
    \newcommand{\imgWidthCelebA}{15.7cm}
    
    \vspace{-5mm}
	\begin{minipage}{10pt}%
    \newcommand{\sizeLeftCelebA}{6.4cm}
	\resizebox{10pt}{!}{\rotatebox{90}{
				\begin{tabular}{ C{\sizeLeftCelebA * 2} C{\sizeLeftCelebA * 2} C{\sizeLeftCelebA * 2} C{\sizeLeftCelebA} C{\sizeLeftCelebA} C{\sizeLeftCelebA} }
					Expand &
					Half
				\end{tabular}%
		}}%
	\end{minipage}%
	\begin{minipage}{\imgWidthCelebA-10pt}%
	\includegraphics[width=\textwidth]{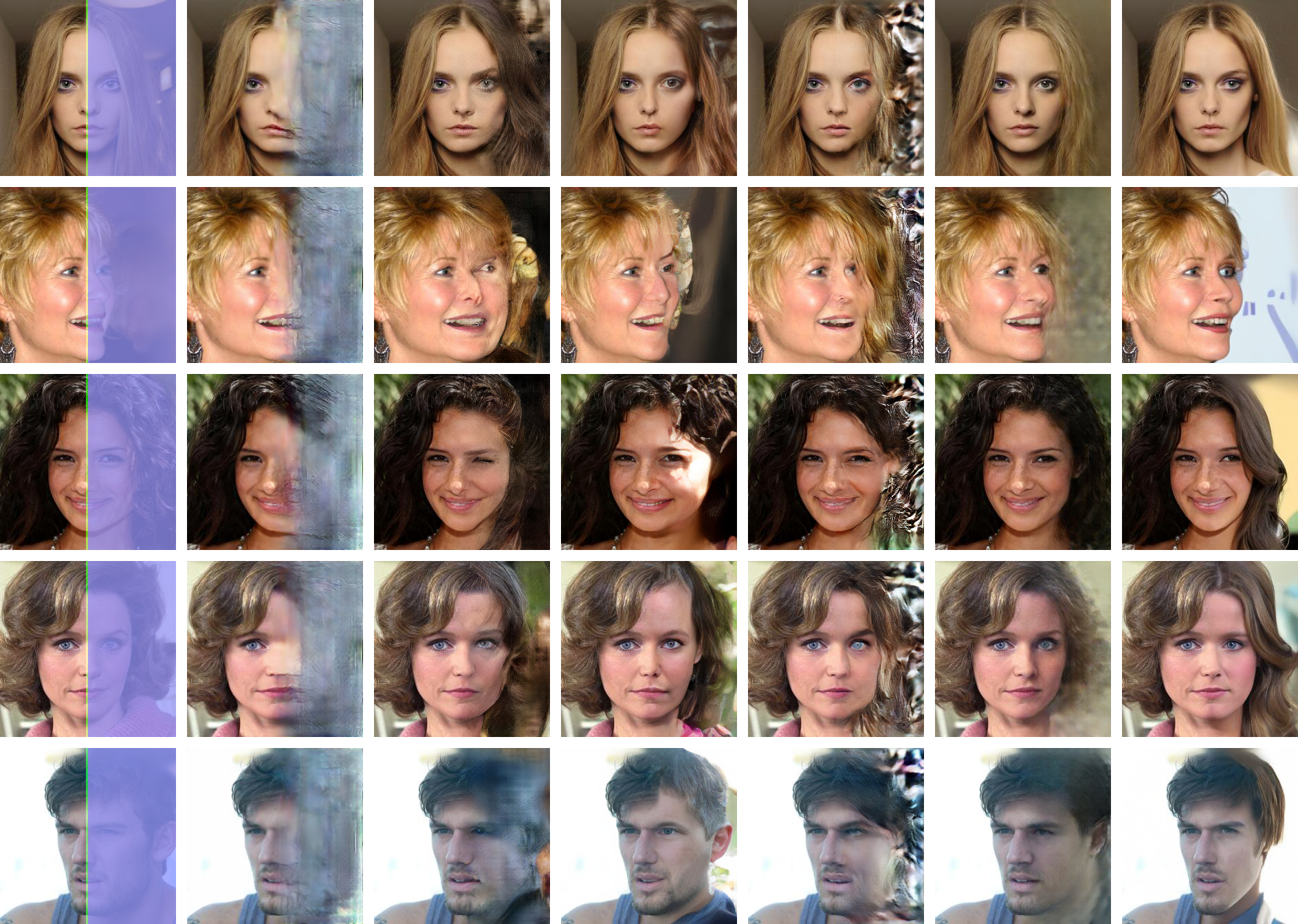} \\
    \vspace{\voffCelebA} \\
	\includegraphics[trim={0 264px 0 0},clip,width=\textwidth]{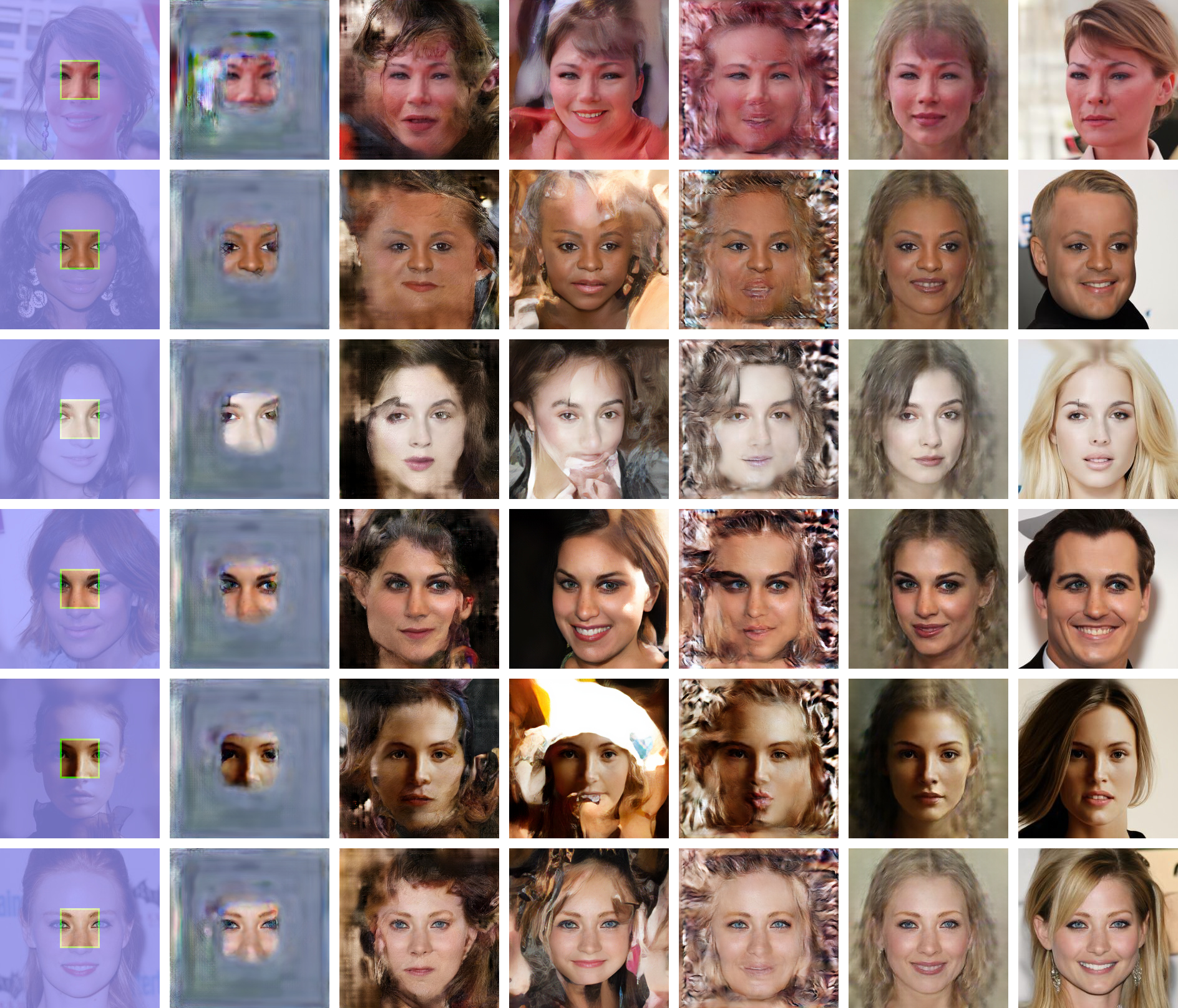} \\
	\end{minipage}
	\begin{minipage}{10pt}%
	\hspace{\fill}
	\end{minipage}%
	\begin{minipage}{{\imgWidthCelebA-10pt}}%
    \captionRowEight{2.5}{
    Input &
    AOT~\cite{aot} &
    DSI~\cite{dsi} &
    ICT~\cite{ict} &
    Deep Fill v2~\cite{deepfillv2} &
    LaMa~\cite{lama} &
    \textbf{RePaint} (ours)
    }
	\end{minipage}%
	\vspace{-0mm}
    \caption{
    \textbf{\dsname Qualitative Results.} Comparison against the state-of-the-art methods for face inpainting. Zoom for better details.}%
    \label{fig:sota_celebA_huge}
\end{figure*}
\def \dsname {ImageNet~}
\def \dstag {inet}

\begin{figure*}
    \centering
    
    \newcommand{\voffCelebA}{-8pt}
    \newcommand{\imgWidthCelebA}{11cm}

	\begin{minipage}{10pt}%
    \newcommand{\sizeLeftCelebA}{6cm}
	\resizebox{10pt}{!}{\rotatebox{90}{
				\begin{tabular}{ C{\sizeLeftCelebA * 2} C{\sizeLeftCelebA * 2} C{\sizeLeftCelebA * 2} C{\sizeLeftCelebA} C{\sizeLeftCelebA} C{\sizeLeftCelebA} }
					Narrow &
					Wide
				\end{tabular}%
		}}%
	\end{minipage}%
	\begin{minipage}{\imgWidthCelebA-10pt}%
	\includegraphics[width=\textwidth]{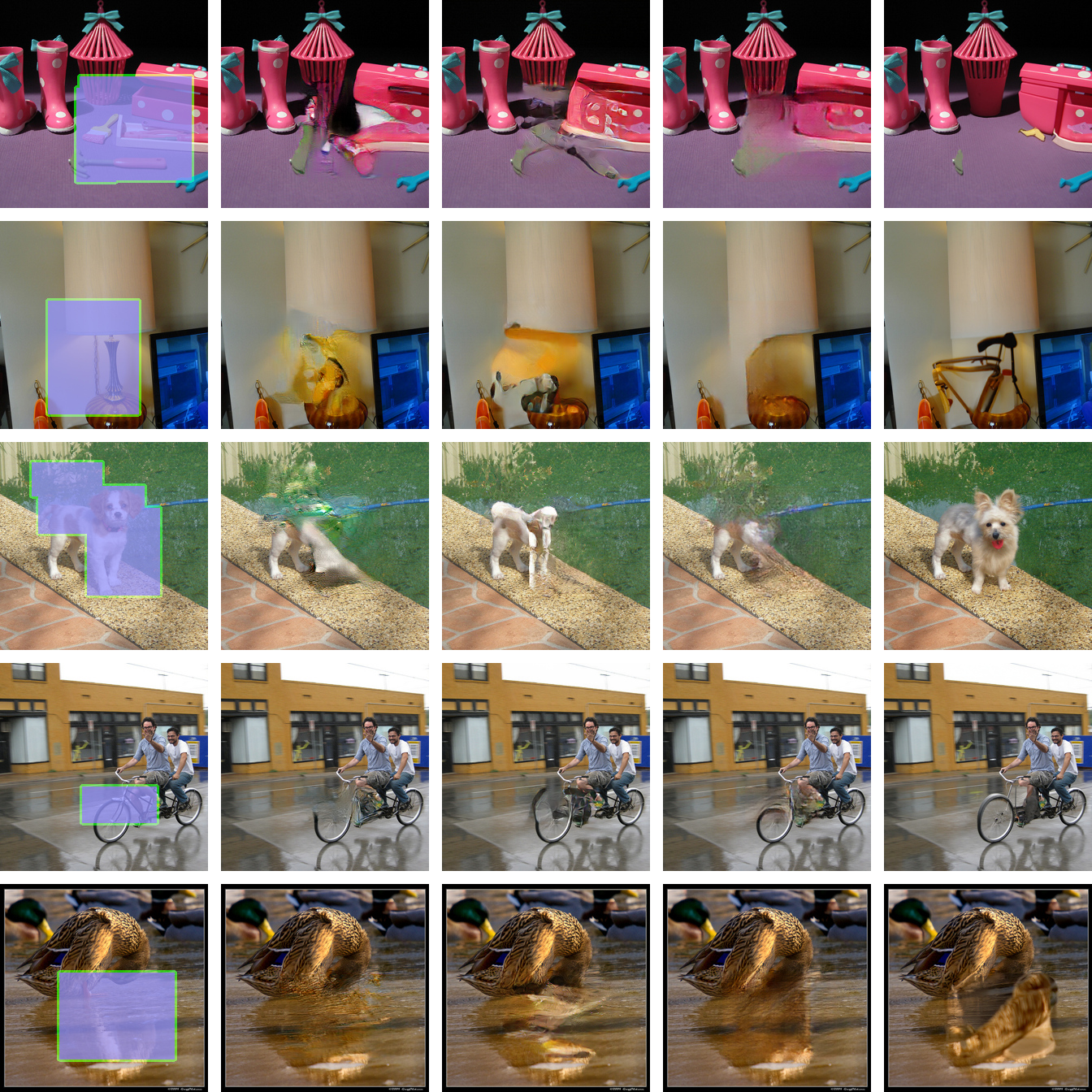} \\
    \vspace{\voffCelebA} \\
	\includegraphics[width=\textwidth]{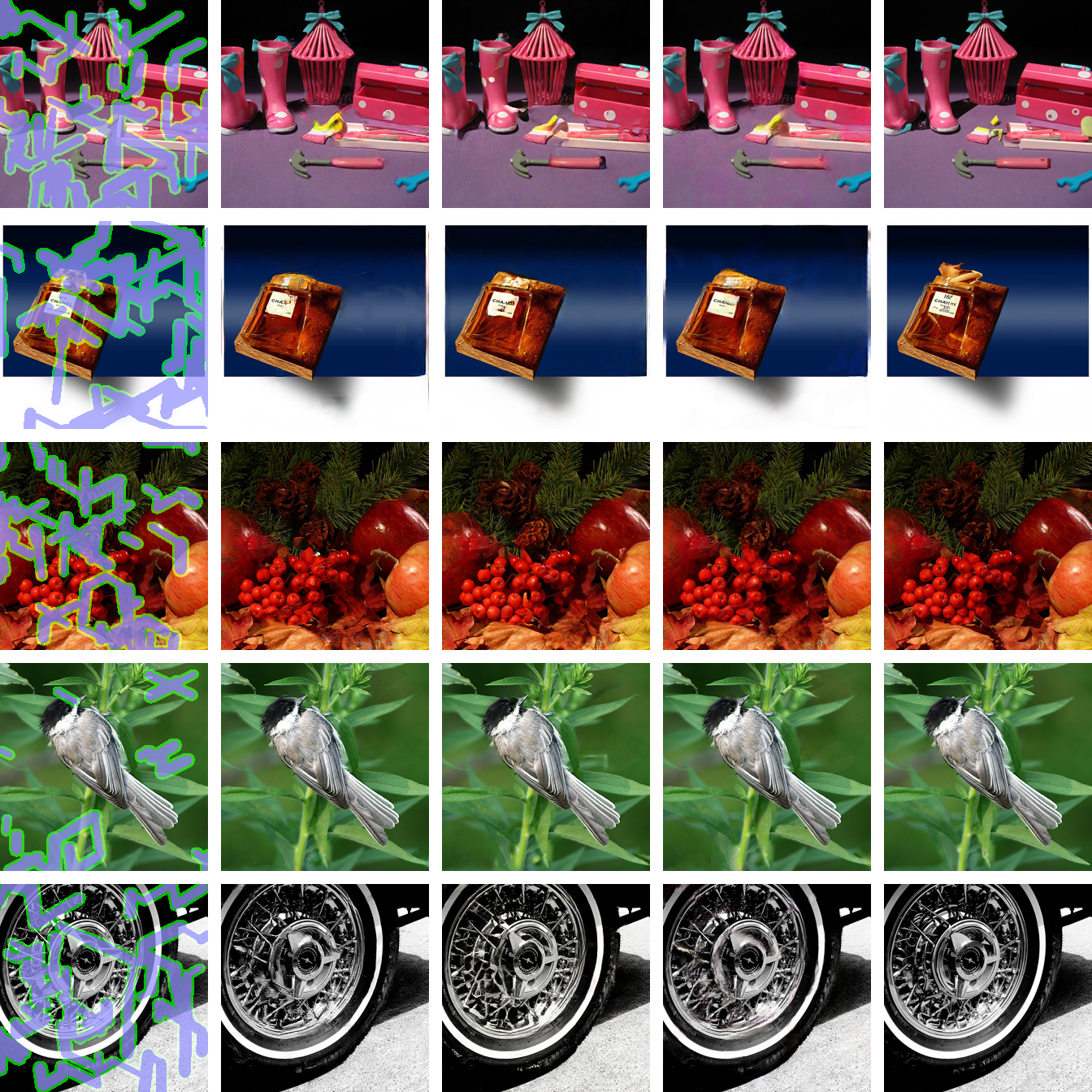} \\
	\end{minipage}
	\begin{minipage}{10pt}%
	\hspace{\fill}
	\end{minipage}%
	\begin{minipage}{{\imgWidthCelebA-10pt}}%
    \captionRowEight{2.5}{
        Input & DSI~\cite{dsi} & ICT~\cite{ict} & LaMa~\cite{lama} & \textbf{RePaint} (ours)
    }
	\end{minipage}%
	\vspace{-0mm}
    \caption{
    \textbf{\dsname Qualitative Results.} Comparison against the state-of-the-art methods for diverse inpainting. Zoom for better details.}%
    \label{fig:sota_inet_lama}
\end{figure*}

\begin{figure*}
    \centering
    
    \newcommand{\voffCelebA}{-8pt}
    \newcommand{\imgWidthCelebA}{11cm}
    
    \vspace{-5mm}
	\begin{minipage}{10pt}%
    \newcommand{\sizeLeftCelebA}{5cm}
	\resizebox{10pt}{!}{\rotatebox{90}{
				\begin{tabular}{ C{\sizeLeftCelebA * 2} C{\sizeLeftCelebA * 2} C{\sizeLeftCelebA * 2} C{\sizeLeftCelebA} C{\sizeLeftCelebA} C{\sizeLeftCelebA} }
					Alternating Lines &
					Super-Resolution $2\times$
				\end{tabular}%
		}}%
	\end{minipage}%
	\begin{minipage}{\imgWidthCelebA-10pt}%
	\includegraphics[width=\textwidth]{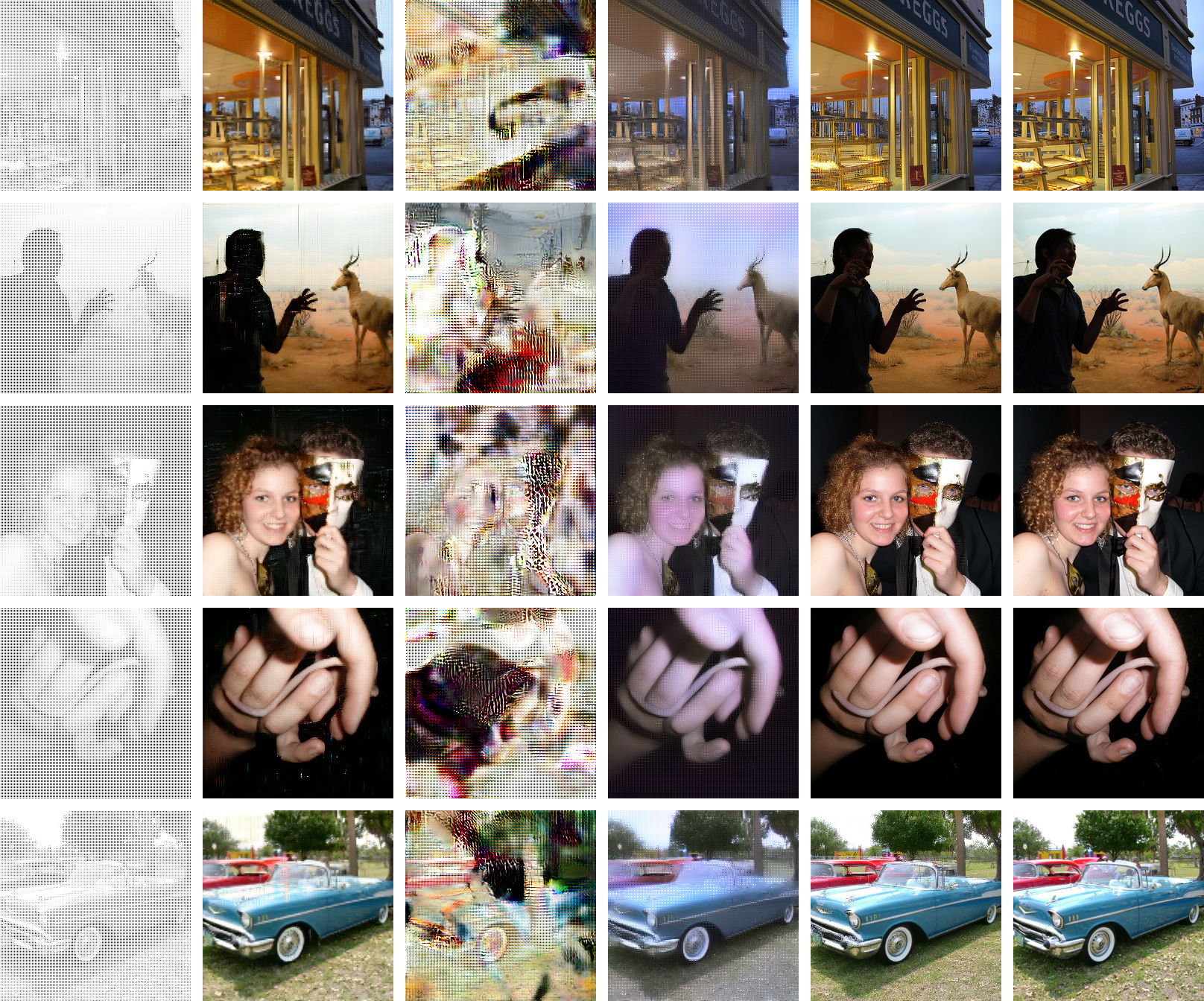} \\
    \vspace{\voffCelebA} \\
	\includegraphics[width=\textwidth]{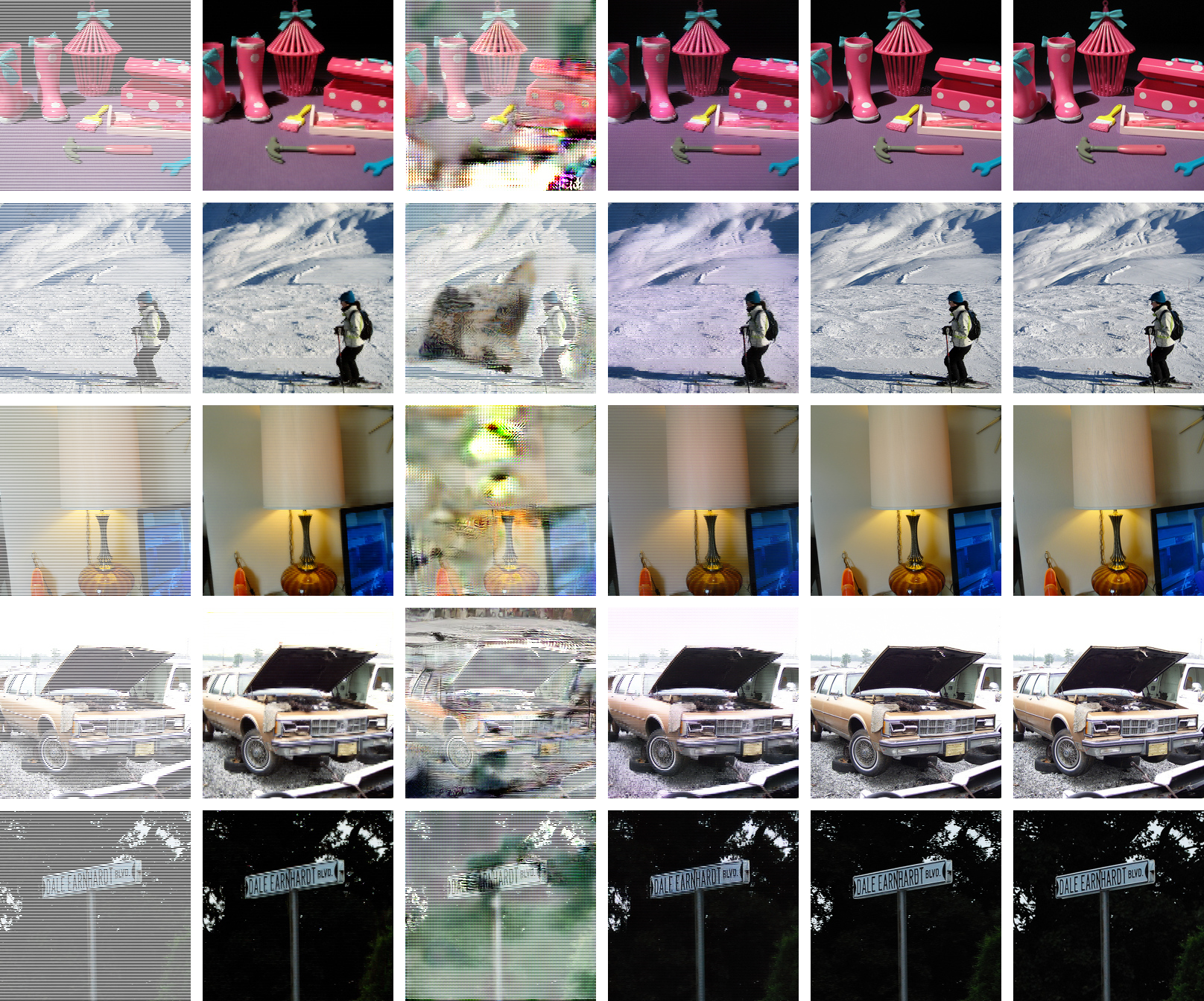} \\
	\end{minipage}
	\begin{minipage}{10pt}%
	\hspace{\fill}
	\end{minipage}%
	\begin{minipage}{{\imgWidthCelebA-10pt}}%
    \captionRowEight{2.5}{
        Input & DSI~\cite{dsi} & ICT~\cite{ict} & LaMa~\cite{lama} &
    \textbf{RePaint} (ours) &
    Ground Truth
    }
	\end{minipage}%
	\vspace{-0mm}
    \caption{
    \textbf{\dsname Qualitative Results.} Comparison against the state-of-the-art methods for diverse inpainting. Zoom for better details.}%
    \label{fig:sota_inet_sparse}
\end{figure*}

\begin{figure*}
    \centering
    
    \newcommand{\voffCelebA}{-8pt}
    \newcommand{\imgWidthCelebA}{11cm}
    
    \vspace{-5mm}
	\begin{minipage}{10pt}%
    \newcommand{\sizeLeftCelebA}{6cm}
	\resizebox{10pt}{!}{\rotatebox{90}{
				\begin{tabular}{ C{\sizeLeftCelebA * 2} C{\sizeLeftCelebA * 2} C{\sizeLeftCelebA * 2} C{\sizeLeftCelebA} C{\sizeLeftCelebA} C{\sizeLeftCelebA} }
					Expand &
					Half
				\end{tabular}%
		}}%
	\end{minipage}%
	\begin{minipage}{\imgWidthCelebA-10pt}%
	\includegraphics[width=\textwidth]{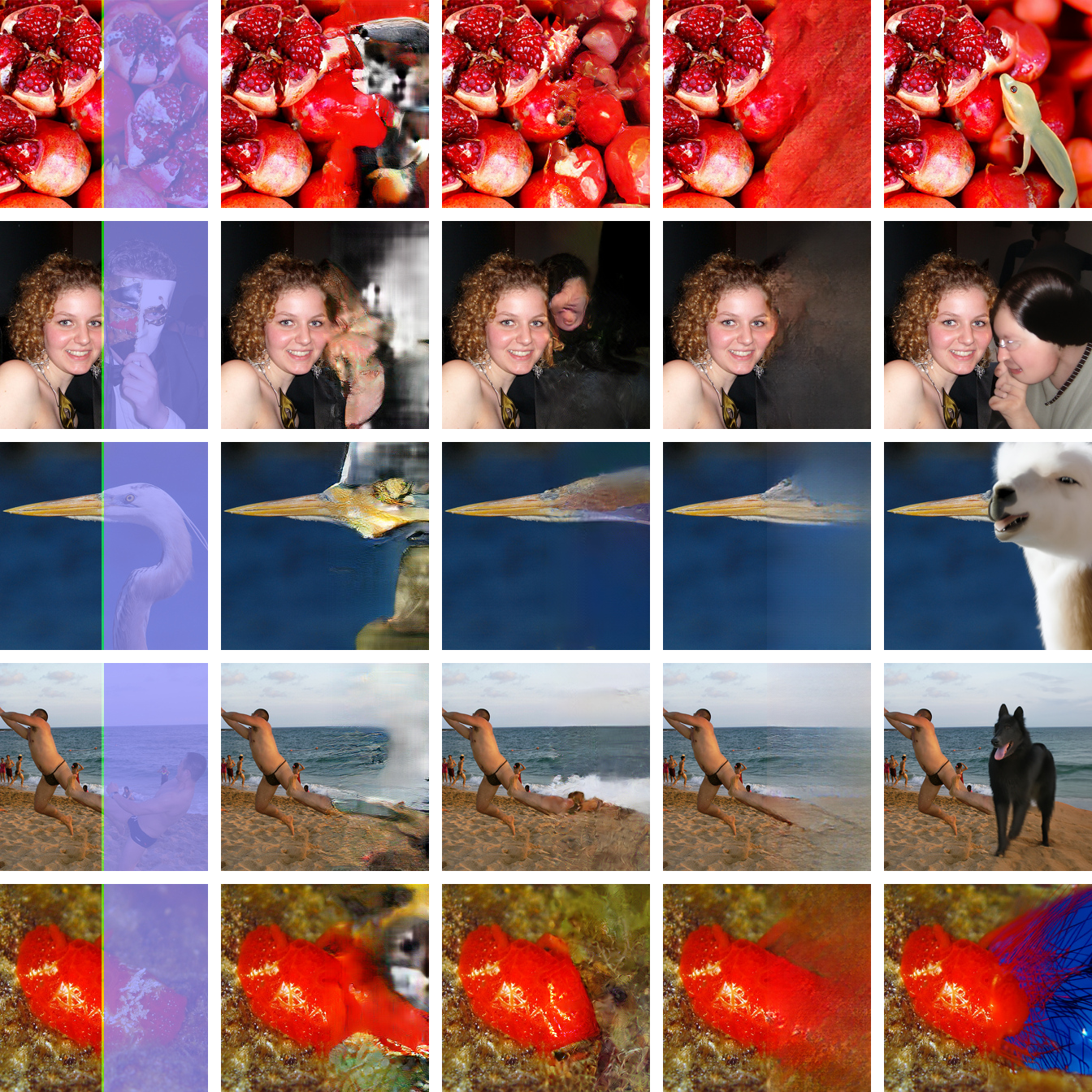} \\
    \vspace{\voffCelebA} \\
	\includegraphics[trim={0 264px 0 0},clip,width=\textwidth]{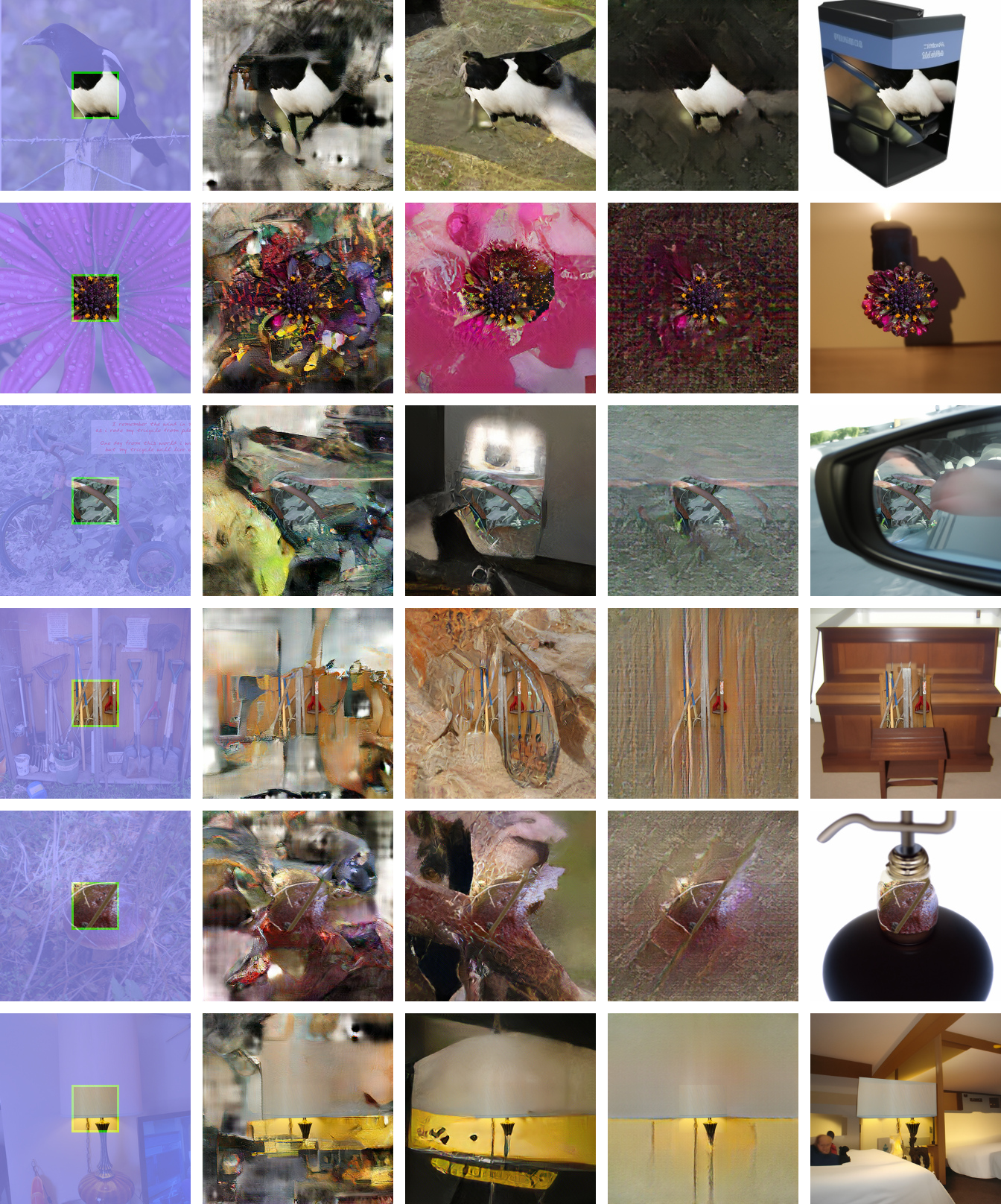} \\
	\end{minipage}
	\begin{minipage}{10pt}%
	\hspace{\fill}
	\end{minipage}%
	\begin{minipage}{{\imgWidthCelebA-10pt}}%
    \captionRowEight{2.5}{
        Input & DSI~\cite{dsi} & ICT~\cite{ict} & LaMa~\cite{lama} & \textbf{RePaint} (ours)
    }
	\end{minipage}%
	\vspace{-0mm}
    \caption{
    \textbf{\dsname Qualitative Results.} Comparison against the state-of-the-art methods for diverse inpainting. Zoom for better details.}%
    \label{fig:sota_inet_huge}
\end{figure*}
\def \dsname {Places2~}
\def \dstag {p}

\begin{figure*}
    \centering
    
    \newcommand{\voffCelebA}{-8pt}
    \newcommand{\imgWidthCelebA}{15.7cm}
    
    \vspace{-5mm}
	\begin{minipage}{10pt}%
    \newcommand{\sizeLeftCelebA}{6.4cm}
	\resizebox{10pt}{!}{\rotatebox{90}{
				\begin{tabular}{ C{\sizeLeftCelebA * 2} C{\sizeLeftCelebA * 2} C{\sizeLeftCelebA * 2} C{\sizeLeftCelebA} C{\sizeLeftCelebA} C{\sizeLeftCelebA} }
					Narrow &
					Wide
				\end{tabular}%
		}}%
	\end{minipage}%
	\begin{minipage}{\imgWidthCelebA-10pt}%
	\includegraphics[width=\textwidth]{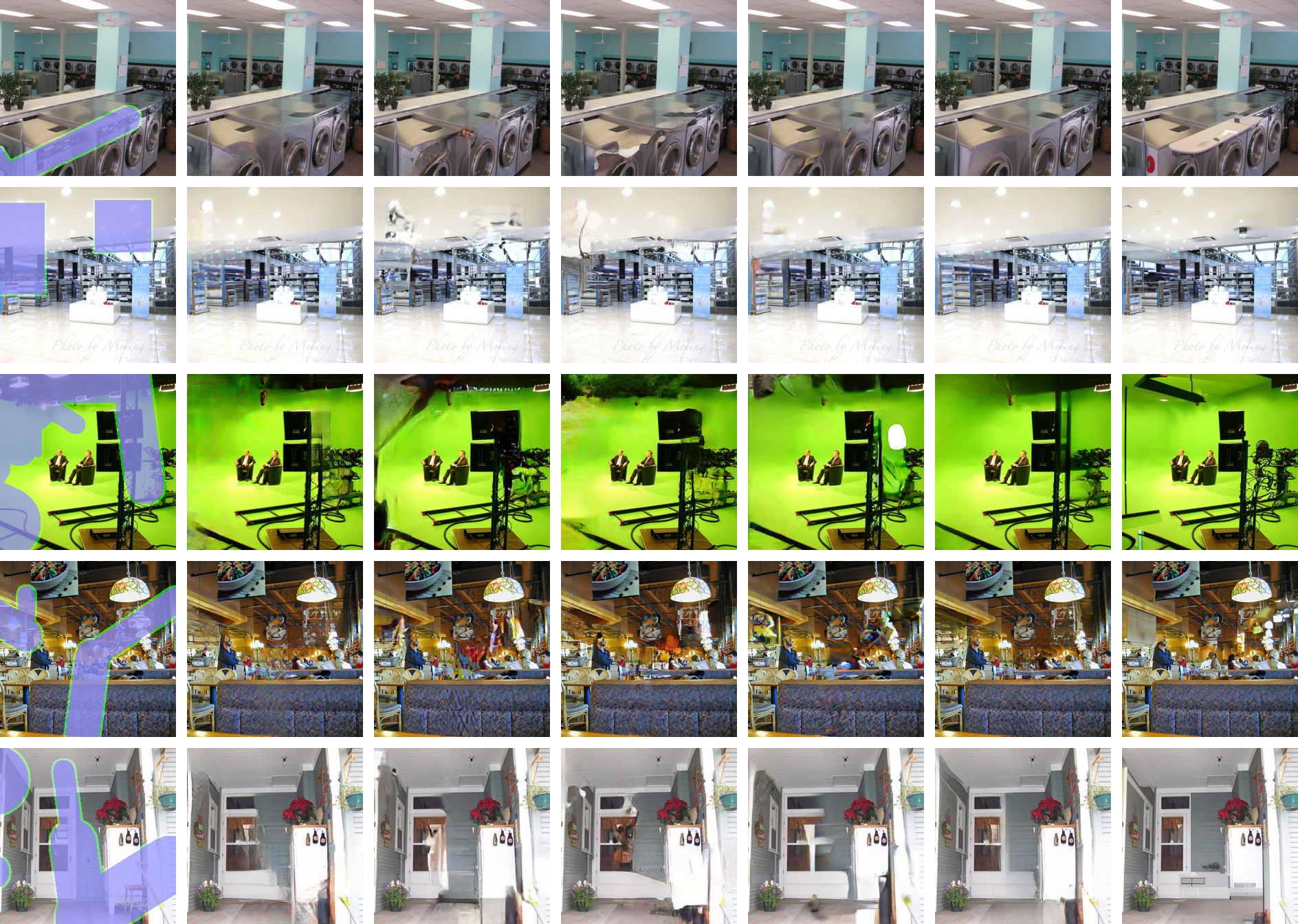} \\
    \vspace{\voffCelebA} \\
	\includegraphics[width=\textwidth]{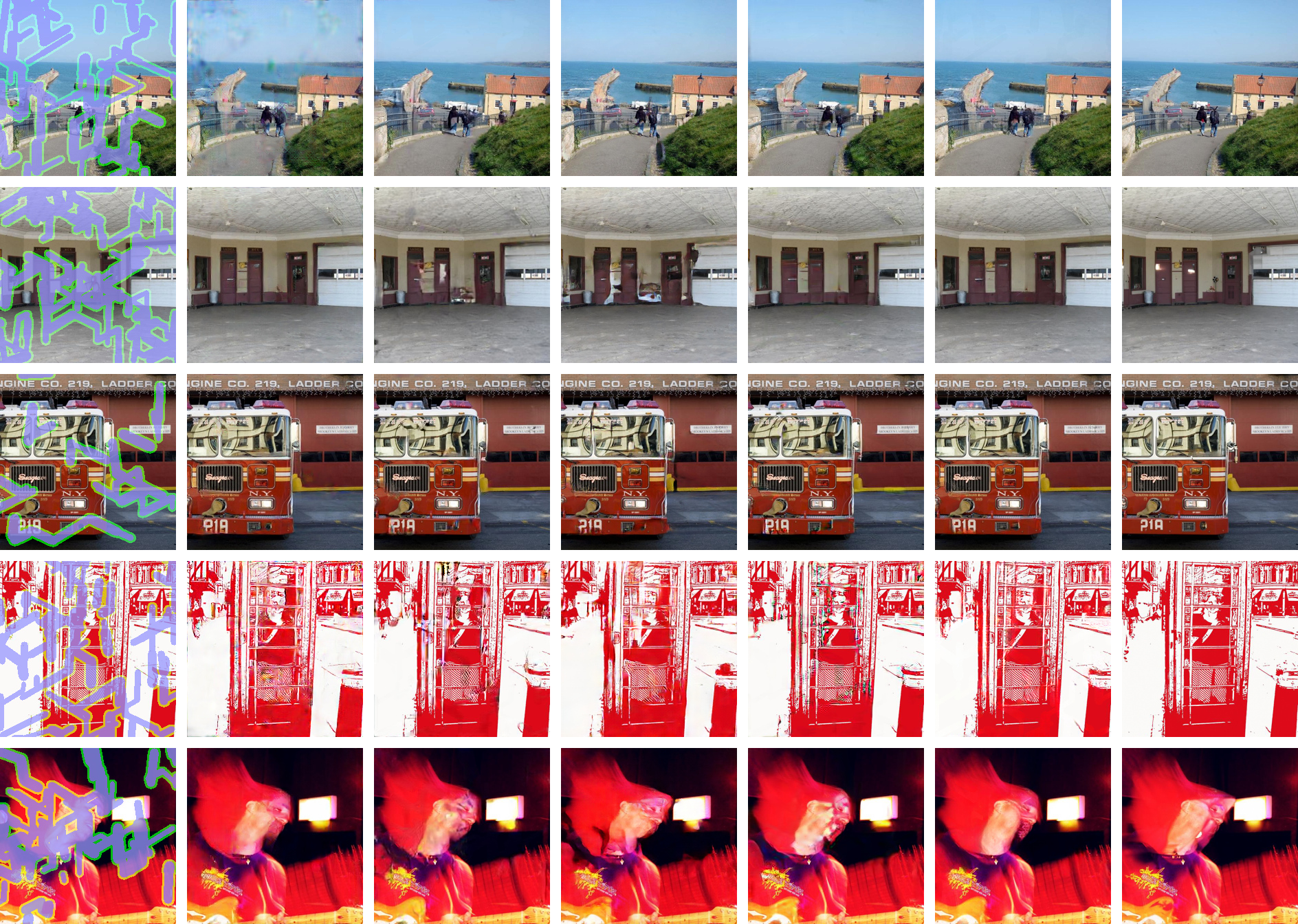} \\
	\end{minipage}
	\begin{minipage}{10pt}%
	\hspace{\fill}
	\end{minipage}%
	\begin{minipage}{{\imgWidthCelebA-10pt}}%
    \captionRowEight{2.5}{
    Input &
    AOT~\cite{aot} &
    DSI~\cite{dsi} &
    ICT~\cite{ict} &
    Deep Fill v2~\cite{deepfillv2} &
    LaMa~\cite{lama} &
    \textbf{RePaint} (ours)
    }
	\end{minipage}%
	\vspace{-0mm}
    \caption{
    \textbf{\dsname Qualitative Results.} Comparison against the state-of-the-art methods for diverse inpainting. Zoom for better details.}%
    \label{fig:sota_places2_lama}
\end{figure*}

\begin{figure*}
    \centering
    
    \newcommand{\voffCelebA}{-8pt}
    \newcommand{\imgWidthCelebA}{15.7cm}
    
    \vspace{-5mm}
	\begin{minipage}{10pt}%
    \newcommand{\sizeLeftCelebA}{5.5cm}
	\resizebox{10pt}{!}{\rotatebox{90}{
				\begin{tabular}{ C{\sizeLeftCelebA * 2} C{\sizeLeftCelebA * 2} C{\sizeLeftCelebA * 2} C{\sizeLeftCelebA} C{\sizeLeftCelebA} C{\sizeLeftCelebA} }
					Alternating Lines &
					Super-Resolution $2\times$
				\end{tabular}%
		}}%
	\end{minipage}%
	\begin{minipage}{\imgWidthCelebA-10pt}%
	\includegraphics[width=\textwidth]{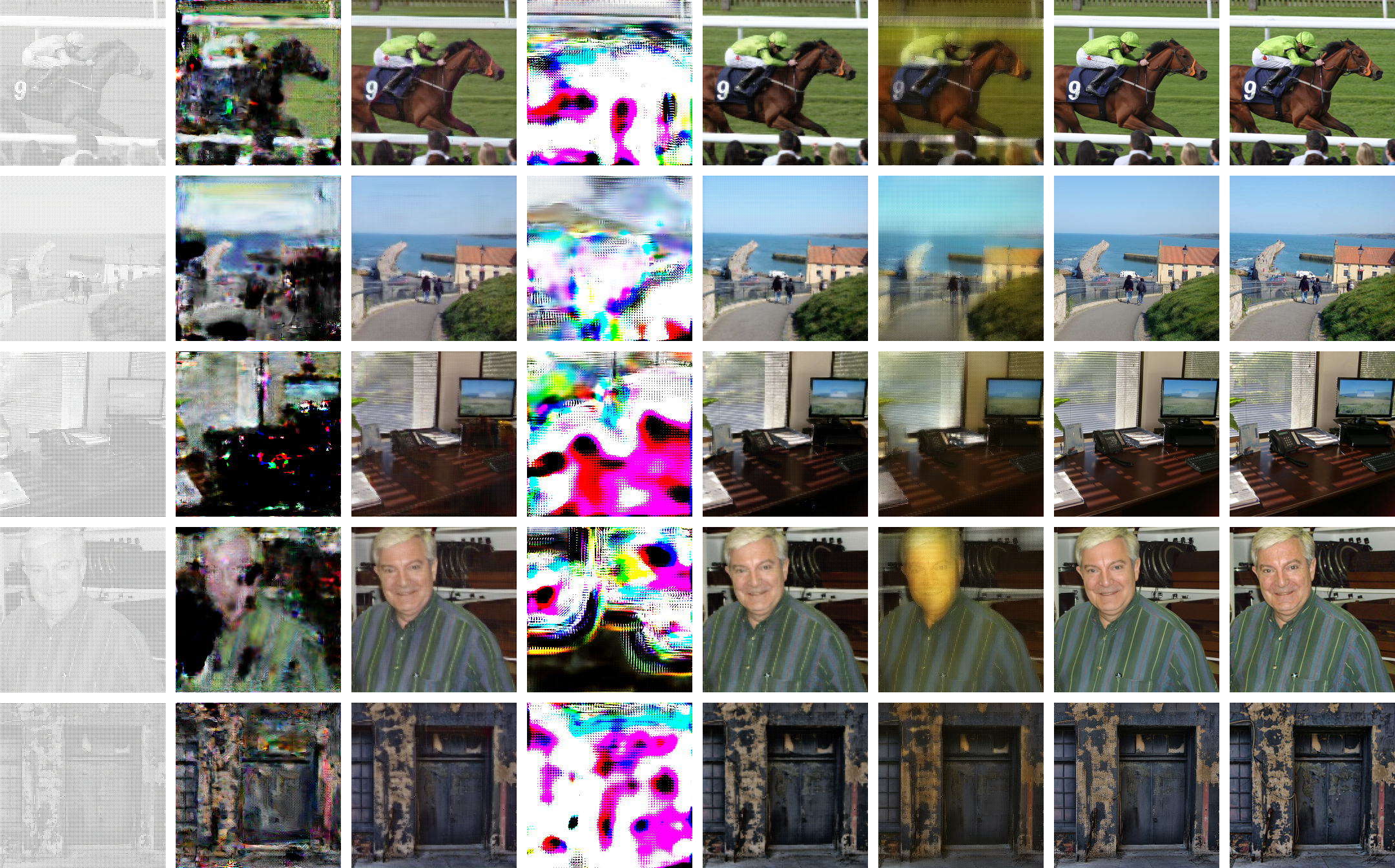} \\
    \vspace{\voffCelebA} \\
	\includegraphics[width=\textwidth]{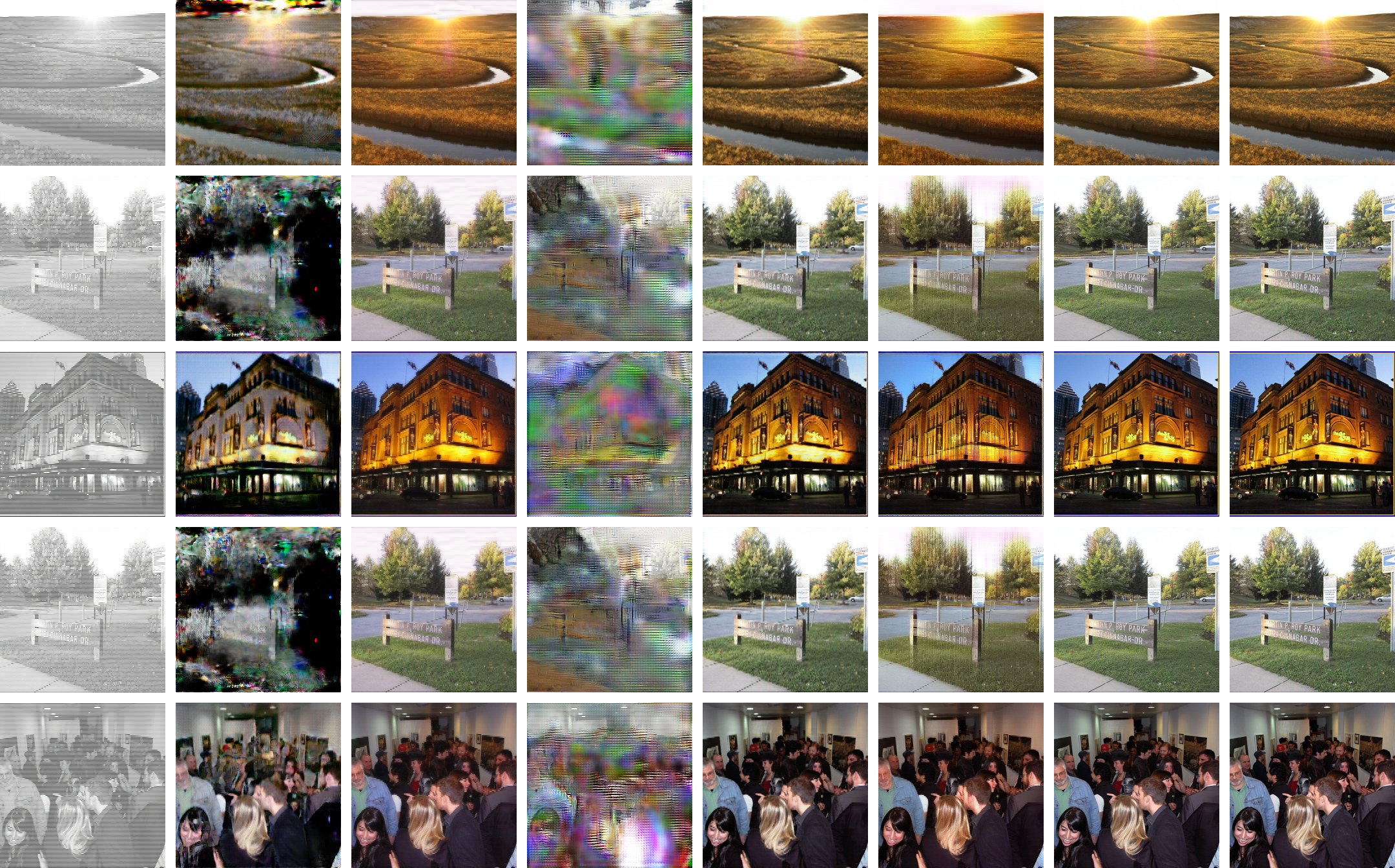} \\
	\end{minipage}
	\begin{minipage}{10pt}%
	\hspace{\fill}
	\end{minipage}%
	\begin{minipage}{{\imgWidthCelebA-10pt}}%
    \captionRowEight{2.5}{
    Input &
    AOT~\cite{aot} &
    DSI~\cite{dsi} &
    ICT~\cite{ict} &
    Deep Fill v2~\cite{deepfillv2} &
    LaMa~\cite{lama} &
    \textbf{RePaint} (ours) &
    Ground Truth
    }
	\end{minipage}%
	\vspace{-0mm}
    \caption{
    \textbf{\dsname Qualitative Results.} Comparison against the state-of-the-art methods for diverse inpainting. Zoom for better details.}%
    \label{fig:sota_places2_sparse}
\end{figure*}

\begin{figure*}
    \centering
    
    \newcommand{\voffCelebA}{-8pt}
    \newcommand{\imgWidthCelebA}{15.7cm}
    
    \vspace{-5mm}
	\begin{minipage}{10pt}%
    \newcommand{\sizeLeftCelebA}{6.4cm}
	\resizebox{10pt}{!}{\rotatebox{90}{
				\begin{tabular}{ C{\sizeLeftCelebA * 2} C{\sizeLeftCelebA * 2} C{\sizeLeftCelebA * 2} C{\sizeLeftCelebA} C{\sizeLeftCelebA} C{\sizeLeftCelebA} }
					Expand &
					Half
				\end{tabular}%
		}}%
	\end{minipage}%
	\begin{minipage}{\imgWidthCelebA-10pt}%
	\includegraphics[trim={0 264px 0 0},clip,width=\textwidth]{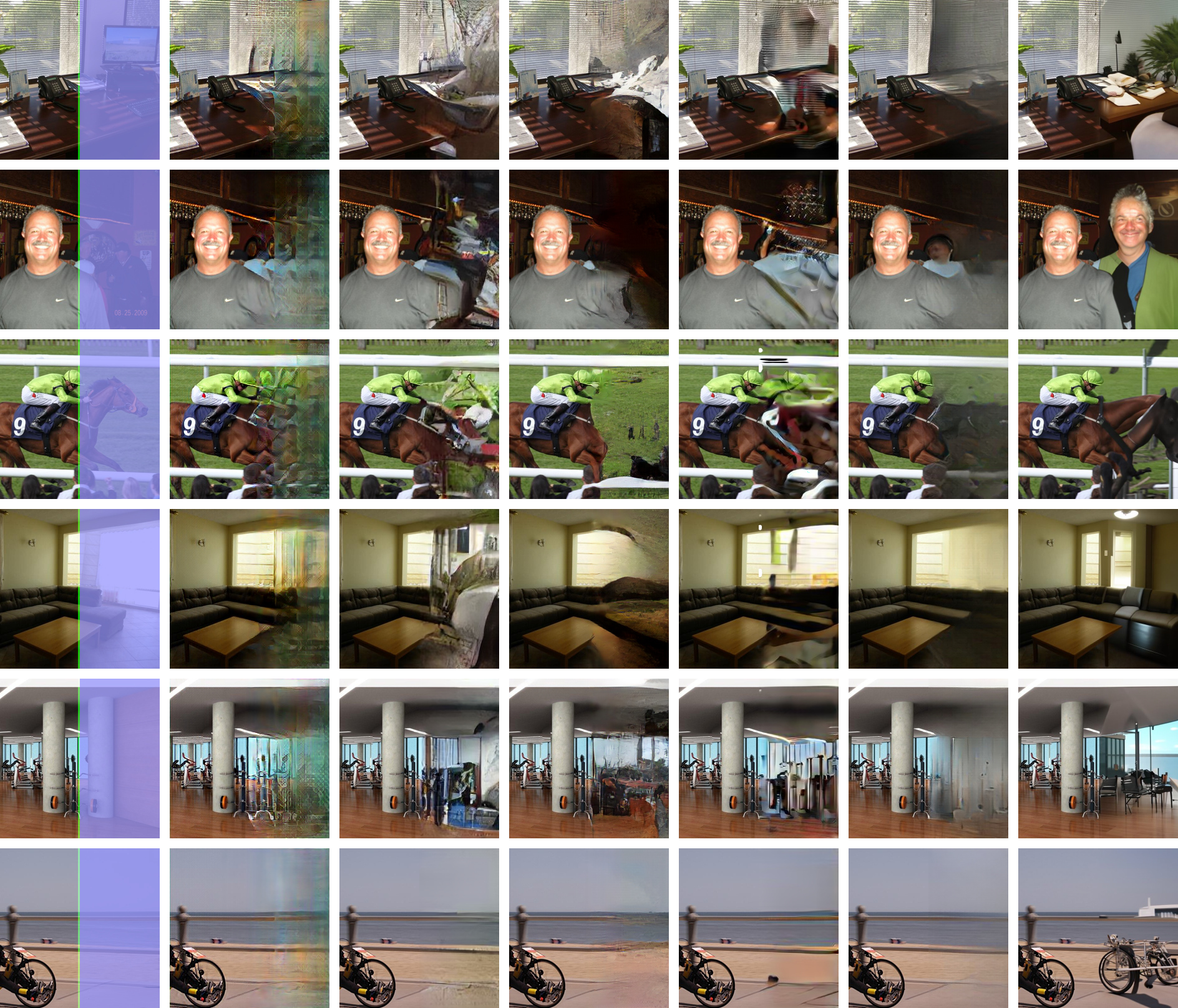} \\
    \vspace{\voffCelebA} \\
	\includegraphics[width=\textwidth]{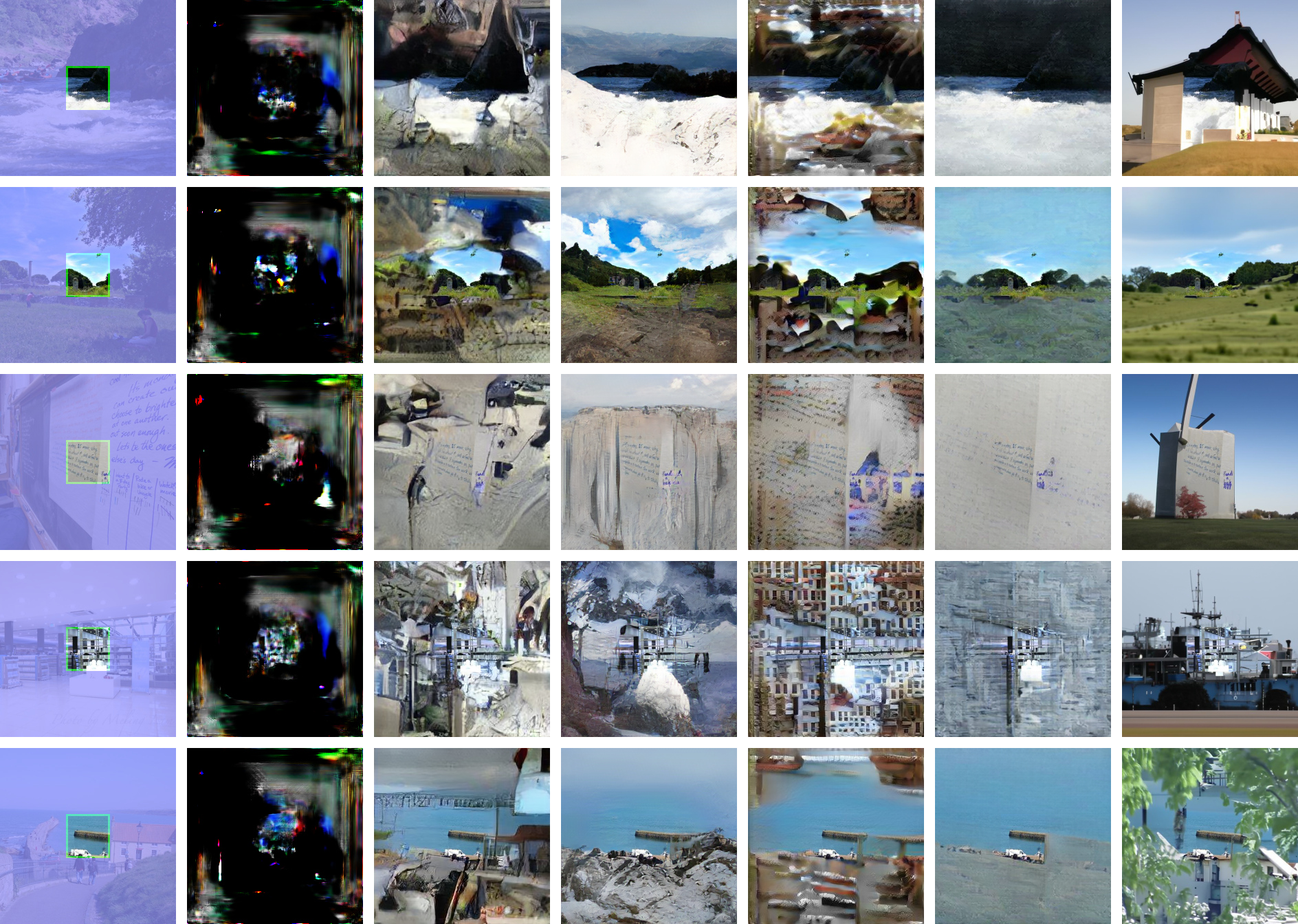} \\
	\end{minipage}
	\begin{minipage}{10pt}%
	\hspace{\fill}
	\end{minipage}%
	\begin{minipage}{{\imgWidthCelebA-10pt}}%
    \captionRowEight{2.5}{
    Input &
    AOT~\cite{aot} &
    DSI~\cite{dsi} &
    ICT~\cite{ict} &
    Deep Fill v2~\cite{deepfillv2} &
    LaMa~\cite{lama} &
    \textbf{RePaint} (ours)
    }
	\end{minipage}%
	\vspace{-0mm}
    \caption{
    \textbf{\dsname Qualitative Results.} Comparison against the state-of-the-art methods for diverse inpainting. Zoom for better details.}%
    \label{fig:sota_places2_huge}
\end{figure*}

\end{document}